\documentclass[sigconf,9pt,table]{acmart}
\AtBeginDocument{%
  \providecommand\BibTeX{{%
    \normalfont B\kern-0.5em{\scshape i\kern-0.25em b}\kern-0.8em\TeX}}}

\acmYear{2024}\copyrightyear{2024}
\acmConference[SenSys '24]{ACM Conference on Embedded Networked Sensor Systems}{November 4--7, 2024}{Hangzhou, China}
\acmBooktitle{ACM Conference on Embedded Networked Sensor Systems (SenSys '24), November 4--7, 2024, Hangzhou, China}
\acmDOI{10.1145/3666025.3699339}
\acmISBN{979-8-4007-0697-4/24/11}

\usepackage{algorithm}
\usepackage{algorithmicx}
\usepackage{algpseudocode}
\usepackage{amsmath,amsfonts}
\usepackage{graphicx}
\usepackage{multirow, makecell}
\usepackage{stfloats}
\usepackage{subfig}
\usepackage{xcolor}
\usepackage{xspace}

\newcommand{\ie}{\emph{i.e.},\xspace}

\newcommand{\eg}{\emph{e.g.},\xspace}
\newcommand{\etc}{\emph{etc.}\xspace}

\newcommand\figref[1]{Fig.~\ref{#1}}

\newcommand\tabref[1]{Tab.~\ref{#1}}

\newcommand\secref[1]{\S~\ref{#1}}

\newcommand\equref[1]{Equ.(\ref{#1})}

\newcommand{\fakeparagraph}[1]{\vspace{1mm}\noindent\textbf{#1.}}

\newcommand{\sysname}{{\sf AdaShadow}\xspace}

\newcommand\lsc[1]{\textcolor{black}{#1}}

\begin{document}
\begin{sloppypar}

\renewcommand{\algorithmicrequire}{\textbf{Input:}}  
\renewcommand{\algorithmicensure}{\textbf{Output:}} 


\title{AdaShadow: Responsive Test-time Model Adaptation in Non-stationary Mobile Environments}



\author{Cheng Fang$^{\dagger}$, Sicong Liu$^{\dagger}$, Zimu Zhou$^{\S}$, Bin Guo$^{\dagger,*}$, Jiaqi Tang$^{\ddagger}$, Ke Ma$^{\dagger}$, Zhiwen Yu$^{\dagger,+}$}
\affiliation{%
  \institution{$^{\dagger}$Northwestern Polytechnical University \\ $^{\S}$City University of Hong Kong \\ $^{\ddagger}$The Hong Kong University of Science and Technology \\ $^{+}$Harbin Engineering University}
  \country{}
}
\thanks{$*$Corresponding email: scliu@nwpu.edu.cn, guob@nwpu.edu.cn}

\begin{abstract}
On-device adapting to continual, unpredictable domain shifts is essential for mobile applications like autonomous driving and augmented reality to deliver seamless user experiences in evolving environments. 
Test-time adaptation (TTA) emerges as a promising solution by tuning model parameters with \textit{unlabeled} live data immediately before prediction. 
However, TTA's unique forward-backward-reforward pipeline notably increases the latency over standard inference, undermining the \textit{responsiveness} in time-sensitive mobile applications.
This paper presents \sysname, a responsive test-time adaptation framework for \textit{non-stationary mobile data distribution} and \textit{resource dynamics} via selective updates of adaptation-critical layers.
Although the tactic is recognized in \textit{generic on-device training}, TTA's \textit{unsupervised} and \textit{online} context presents unique challenges in estimating layer importance and latency, as well as scheduling the optimal layer update plan.
\sysname addresses these challenges with a \textit{backpropagation-free assessor} to rapidly identify critical layers, a unit-based \textit{runtime predictor} to account for resource dynamics in latency estimation, and an \textit{online scheduler} for prompt layer update planning. 
Also, \sysname incorporates a memory I/O-aware computation reuse scheme to further reduce latency in the reforward pass.
Results show that \sysname achieves the best accuracy-latency balance under continual shifts. 
At low memory and energy costs, Adashadow provides a $2\times$ to $3.5\times$ speedup (\textit{ms-}level) over state-of-the-art TTA methods with comparable accuracy and a 14.8\% to 25.4\% accuracy boost over efficient supervised methods with similar latency.

\end{abstract}



\maketitle

\section{Introduction}
\label{sec:intro}

Deep neural networks (DNNs) pre-trained in the cloud are increasingly deployed onto mobile devices for autonomous intelligence at the edge \cite{liu2023enabling, he2023vi, tang2023lut, liu2021adaspring, wang2023adaevo, liu2023adaenlight}, especially in some cities where cloud streaming is restricted~\cite{AzureDataResidency, SwedenDataResidency}.
Applications such as on-device motion tracking (\eg Google ARcore~\cite{google}) and AR/VR (\eg Apple ARKit~\cite{apple}, Meta Spark~\cite{meta}).
\lsc{
These DNNs must operate reliably in open-world mobile environments, necessitating \textit{swift model adaptation to unforeseen domain shifts} which represent differences in the distribution of the pre-trained model's data and the data encountered during test. 
The shifts are mainly caused by \textit{environmental changes} (e.g., weather, lighting) and \textit{sensor degradation} (e.g., low resolution and Gaussian noise).
}
For instance, an autonomous car may encounter temporary road signs or modified traffic signals due to construction work and must adjust its DNNs with a few live samples collected on-the-fly \cite{wang2018networking, liu2020computing}. 
Such adaptation is crucial for maintaining safety and operational efficiency. 
Similarly, AR headsets require seamless integration of virtual elements into the physical world. 
As conditions such as lighting, field of view, and backgrounds vary with user movement, these headsets need to \textit{promptly} refine their object recognition/tracking models based on limited input video clips, keeping the virtual overlays consistent with the evolving physical surroundings \cite{jain2015overlay,corbett2023bystandar,xu2024improving}.
For instance, a vehicle driving at 90km/h requires its 60fps traffic sign recognition model to adapt within 16.6ms to ensure safety.

Test-time adaptation (TTA) offers a compelling solution to combat \textit{unpredictable domain shifts} in mobile environments \cite{liang2023comprehensive}. 
It is an emerging domain adaptation paradigm that \textit{(i)} utilizes unlabeled test data, and \textit{(ii)} operates independently of source data and supervision from pre-training \cite{wang2021tent}.
TTA typically works in three stages.
A \textit{forward} pass first generates initial inference results on the shifted test samples, \ie the live data sensed in a new environment. 
If inference confidence is low (\eg high entropy), then a \textit{backward} pass follows to adjust the DNN parameters using unsupervised loss based on initial forward results.
Afterward, a \textit{reforward} pass applies the updated DNN to the same batch of input data to make the final predictions. 
This continual, \textit{unsupervised}, \textit{source-free} adaptation is particularly advantageous for mobile applications, which often lack access to pre-training data due to privacy or network restrictions. 
Moreover, these applications typically preclude the possibility of pre-labeling target domains due to the requisite labour and resources.

Despite advances in TTA algorithms \cite{wang2021tent, wang2022continual, hong2023mecta}, their practical adoption in mobile applications is challenging. 
The \textit{forward-backward-reforward} pipeline of TTA, essential for adaptability to domain shifts, adds considerable \textit{latency} compared to standard inference (see \secref{sec:background:latency:inefficiency}), which compromises the \textit{responsiveness} in \textit{time-sensitive} mobile applications \eg AR and autonomous cars, where even minor delays can drastically impact user experience and operational safety. 
Despite pioneer studies on efficient TTA \cite{niu2022efficient, hong2023mecta, song2023ecotta}, they focus on decreasing computational or memory cost, which cannot easily translate into reduced \textit{wall-clock time}.
Backward-free TTA  \cite{wang2024backpropagation,wang2024optimization,boudiaf2022parameter} experience unacceptable accuracy degradation during continual adaptation.
This gap necessitates \textit{latency-efficient} TTA for \textit{fast} and \textit{accurate} adaptation to \lsc{non-stationary environments for mobile computing}.

In this paper, we propose \sysname, a novel framework for \textit{responsive} test-time adaptation, like how a shadow \textit{swiftly} responds to the body's movements.
The principle is \textit{sparse updating}, \ie selectively refining critical layers to reduce both computation and memory demands during backpropagation, and thus latency \cite{zaken2022bitfit, mudrakarta2019k, huang2023elastictrainer}.
Although the tactic has been explored to improve the latency efficiency of generic DNN training on mobile devices \cite{huang2023elastictrainer}, sparse updating in TTA encounters unique challenges due to its \textit{unsupervised} and \textit{online} nature. We elaborate on these challenges below.

\noindent$\bullet$ \textbf{Challenge $\#1$: identifying critical layers at low latency}. The importance of each layer differs drastically across different data distributions \cite{lee2022surgical}. 
In \lsc{non-stationary environments for mobile computing}, each data batch may exhibit unique domain shifts, demanding rapid layer importance assessment. 
Prior works~\cite{huang2023elastictrainer, lin2022device} assess layer importance by computing the gradients using \textit{labeled} data, which are computation-intensive (\ie higher latency) and error-prone in TTA as test data are usually unlabeled in highly dynamic mobile scenarios.
For instance, ElasticTrainer~\cite{huang2023elastictrainer} results in $2.98s$ delay and 14.8\% gradient errors for ResNet50 on unlabeled data.

\noindent$\bullet$ \textbf{Challenge $\#2$: predicting runtime latency for layer update}. To ensure the updated model architecture meets the latency requirement, we have to know exactly the run-time latency of each retained layer in this new model.
Prior works~\cite{zhang2021nn, huang2023elastictrainer} measure layer latency offline and then use it to predict the run-time model inference latency. Yet mobiles are interactive devices, their hardware resource availability will change dynamically due to user oprations, foreground-background APP switching, and computation and memory resource competitions, which makes offline estimation inaccurate for online model inference prediction.


\noindent$\bullet$  \textbf{Challenge $\#3$: online scheduling layer update strategy efficiently}. 
Even with accurate information on each layer's importance and their runtime latency, finding the optimal sparse updating strategy to improve performance in the presence of rigid delay requirements is still challenging due to the large search space. 
Dedicated strategies are necessary in exploring and pruning the search space with high efficiency.

\begin{table}[t]
\footnotesize
\caption{Differences of our work from related schemes.}
\renewcommand{\arraystretch}{1.05}
\scalebox{1}{
\begin{tabular}{|cl|c|c|c|c|}
\hline
\multicolumn{2}{|c|}{\multirow{2}{*}{\textbf{Method}}} & \multirow{2}{*}{\textbf{\begin{tabular}[c]{@{}c@{}}Support \\ unlabeled data\end{tabular}}} & \multirow{2}{*}{\textbf{\begin{tabular}[c]{@{}c@{}}Online\\ adaptation\end{tabular}}} & \multirow{2}{*}{\textbf{\begin{tabular}[c]{@{}c@{}}Low\\ latency\end{tabular}}} & \multirow{2}{*}{\textbf{\begin{tabular}[c]{@{}c@{}}High \\ accuracy\end{tabular}}} \\
\multicolumn{2}{|c|}{} &  &  &  &  \\ \hline
\multicolumn{2}{|c|}{\begin{tabular}[c]{@{}c@{}}Efficient training: \\ Melon \cite{wang2022melon}\end{tabular}} & No & No & No & Yes \\ \hline
\multicolumn{2}{|c|}{\begin{tabular}[c]{@{}c@{}}Efficient training: \\ ElasticTrainer \cite{huang2023elastictrainer}\end{tabular}} & No & No & Yes & Yes \\ \hline
\multicolumn{2}{|c|}{\begin{tabular}[c]{@{}c@{}}Generic TTA: \\ CoTTA \cite{wang2022continual}\end{tabular}} & Yes & Yes & No & Yes \\ \hline
\multicolumn{2}{|c|}{\begin{tabular}[c]{@{}c@{}}Efficient TTA: \\ EcoTTA \cite{song2023ecotta}\end{tabular}} & Yes & Yes & No & Yes \\ \hline
\multicolumn{2}{|c|}{\begin{tabular}[c]{@{}c@{}}Efficient TTA:\\  LAME \cite{boudiaf2022parameter}\end{tabular}} & Yes & No & Yes & No \\ \hline
\multicolumn{2}{|c|}{\sysname} & Yes & Yes & Yes & Yes \\ \hline
\end{tabular}
}
\label{tb:comparison}
\end{table}

\sysname addresses these challenges with three functional module designs.

\noindent \textbf{First}, we notice that the forward pass executed with each inference in TTA offers an opportunity to evaluate layer importance, avoiding the higher latency and imprecise results of backward gradients. 
Based on this observation, we propose a backpropagation-free layer importance assessor that can timely assess the importance of each layer by measuring the \textit{divergence} between the layers' \textit{output feature maps} in different environments, without requiring labeled data.

\noindent \textbf{Second}, 
to deliver precise and timely adaptation latency feedback, we introduce an online, unit-based (\ie layer) adaptation latency predictor. 
This predictor distinguishes between static and dynamic update delays and incorporates fine-grained, dynamic system metrics like cache hit rates, competing CPU/GPU processes, and frequency into each unit's latency for accurate runtime measurements and improved online prediction.

\noindent \textbf{Third}, 
Given predictions of unit importance and update latency, we further develop a lightweight dynamic programming (DP)-based online scheduler. 
This scheduler efficiently determines the optimal layer update strategy for three-stage TTA by clearly defining specific subproblems, utilizing recursion, and eliminating invalid subproblems.

In the implementation, \sysname optimizes the TTA loss for \textit{small batches}, which is common in mobile applications, and harnesses \textit{computation reuse} between forward and reforward passes to boost performance.
We evaluate \sysname across three real-world mobile scenarios, addressing over 30 types of data shifts and 5 types of resource dynamics, using three mobile devices with varied hardware architectures. 
\lsc{
Results show that \sysname achieves the best accuracy-latency balance in scenarios with continual shifts.
Adashadow achieves a 2× to 3.5× adaptation speedup compared to state-of-the-art TTA methods, while maintaining comparable accuracy. 
Additionally, it offers a 14.8\% to 25.4\% accuracy improvement over state-of-the-art efficient supervised adaptation methods with similar latency   (\secref{sec:exp}).
}
Our main contributions are summarized as follows.
\begin{itemize}
\item 
To our knowledge, this is the first work on near-/real-time on-device DNN adaptation in \lsc{non-stationary environments for mobile computing} without labels or source data access. 
It overcomes latency bottlenecks of TTA in mobile contexts without compromising accuracy.
\item 
We propose \sysname, a holistic system design that includes a backpropagation-free layer importance assessor, a runtime latency predictor, an online update scheduler, and optimizations for efficient, resource-aware TTA. 
It seamlessly integrates with mainstream TTA pipelines, supports various DNN architectures.
\item 
Experiments show that \sysname outperforms existing research on generic TTA \cite{wang2021tent, wang2022continual}, efficient TTA \cite{niu2022efficient, song2023ecotta, hong2023mecta}, and efficient training (with labels) \cite{wang2022melon, lin2022device, huang2023elastictrainer} in trading-off accuracy and latency at low memory and energy costs over diverse tasks, shifts, and devices. 
\end{itemize}

\section{Background}
\label{sec:background}

\subsection{Primer on Test-time Adaptation (TTA)}
\label{sec:background:why}

Test-time adaptation (TTA) is an emerging \textit{unsupervised domain adaptation} paradigm to combat \textit{domain shifts} \cite{liang2023comprehensive}.
It adapts the DNN \textit{pre-trained} in the source domain to \textit{unlabelled} target data during \textit{testing}, enhancing the accuracy on target data \cite{wang2021tent}.
Uniquely, TTA assumes no access to the \textit{source data} and \textit{supervision} from the pre-training stage.
This setup is fit for mobile applications where \textit{(i)} accessing source data is impractical due to privacy concerns or bandwidth limitations; and \textit{(ii)} annotating the target domain is infeasible or labor-intensive. 
However, TTA often introduces significant latency (see details in \secref{sec:background:latency}) or decline in accuracy if overfitting to the new environment~\cite{wang2022continual,niu2022efficient}.

We explore \textit{responsive} test-time adaptation of DNNs to \textit{non-stationary} mobile domain shifts, with a focus on \textbf{latency efficiency}.
This is because many mobile applications demand \textit{real-time responsiveness}.
Accordingly, the DNN should adapt swiftly, delivering accurate inference under continuous domain shifts without perceptible delays. 
Additionally, the non-stationary environment means the adaptation should \textit{(i)} operate effectively with \textit{small batches} of data; and \textit{(ii)} account for \textit{runtime resources dynamics}. 
Both requirements impose extra challenges when optimizing the latency of TTA (see \secref{sec:intro}).
\tabref{tb:comparison} summarizes the differences of our work from other representative studies.

\subsection{Latency of TTA}
\label{sec:background:latency}

\textbf{Inefficiency of Prior Arts}.
\label{sec:background:latency:inefficiency}
Despite emerging research on test-time adaptation for improved data efficiency \cite{niu2022efficient} and memory efficiency \cite{hong2023mecta,song2023ecotta}, latency remains a less-explored challenge. 
From \tabref{tb:tta_latency}, EcoTTA \cite{song2023ecotta}, a state-of-the-art, can be up to $4.1\times$ slower than inference using the pre-trained model without adaptation (denoted as \textit{source} in the table), even though they achieve higher accuracy on the drifted testing datasets.
The unsatisfactory latency motivates us to zoom into the latency of TTA.

\begin{table}[t]
\centering
\footnotesize
\caption{Accuracy and latency of state-of-the-art TTA methods against inference w/o adaptation.}
\renewcommand{\arraystretch}{1.1}
\scalebox{0.95}{
\begin{tabular}{|c|ccc|ccc|}
\hline
 & \multicolumn{3}{c|}{NICO++} & \multicolumn{3}{c|}{CIFAR10-C} \\ \hline
Method & \multicolumn{1}{c|}{ACC.(\%)} & \multicolumn{1}{c|}{\begin{tabular}[c]{@{}c@{}}Latency\\ (ms)\end{tabular}} & Th.(fps) & \multicolumn{1}{c|}{ACC.(\%)} & \multicolumn{1}{c|}{\begin{tabular}[c]{@{}c@{}}Latency\\ (ms)\end{tabular}} & Th.(fps) \\ \hline
Source & \multicolumn{1}{c|}{60.7} & \multicolumn{1}{c|}{10.2} & 98.2 & \multicolumn{1}{c|}{52.2} & \multicolumn{1}{c|}{9.8} & 102.2 \\ \hline
Tent\cite{wang2021tent} & \multicolumn{1}{c|}{87.8} & \multicolumn{1}{c|}{33.2} & 30.1 & \multicolumn{1}{c|}{77.8} & \multicolumn{1}{c|}{25.8} & 38.7 \\ \hline
EATA\cite{huang2023elastictrainer} & \multicolumn{1}{c|}{88.4} & \multicolumn{1}{c|}{36.5} & 27.4 & \multicolumn{1}{c|}{79.2} & \multicolumn{1}{c|}{27.5} & 36.4 \\ \hline
EcoTTA\cite{song2023ecotta} & \multicolumn{1}{c|}{89.2} & \multicolumn{1}{c|}{42} & 23.8 & \multicolumn{1}{c|}{80.1} & \multicolumn{1}{c|}{31.1} & 32.2 \\ \hline
\end{tabular}
}
\label{tb:tta_latency}
\end{table}

\textbf{Latency Bottleneck}.
\label{sec:background:latency:bottleneck}
Latency is related to computing, memory access, and the availability of hardware resources~\cite{zhang2021nn, kong2023convrelu++}.
A typical TTA pipeline \cite{liang2023comprehensive} includes three phases: \textit{forward}, \textit{backward}, and \textit{reforward}. 
In the first two phases (forward and backward), the model updates its parameters via standard gradient descent, based on the input batch data. 
In the last phase (reforward), the model performs inference on the \textit{same} batch of data utilizing freshly updated parameters. 
This pipeline, known as the \textit{sequential adaptation/inference} mode\footnote{A few studies \cite{wang2021tent, niu2022efficient} prioritize latency over accuracy in process scheduling by only applying the newly adapted model to subsequent sample inference, thus eliminating the reforward latency.
Our method achieves higher accuracy with lower latency than this execution mode (see \secref{sec:exp:case}).}, prioritizes inference accuracy by adapting the model before making predictions on it.
However, it also implies that the inference for any batch of data must wait for the completion of adaptation, leading to noticeable latency.
Formally, the overall TTA latency can be calculated as:
\begin{equation} \label{d1}
T=T_{a}+T_{re}=T_{f}+T_{b}+T_{re}
\end{equation}
where $T_{a}$ and $T_{re}$ are the latency of model adaptation and inference (reforward), and $T_{a}$ is further decomposed into $T_{f}$ and $T_{b}$, \ie latency of forward and backward, respectively.

\begin{figure}[t]
  \centering
  \includegraphics[width=0.75\linewidth]{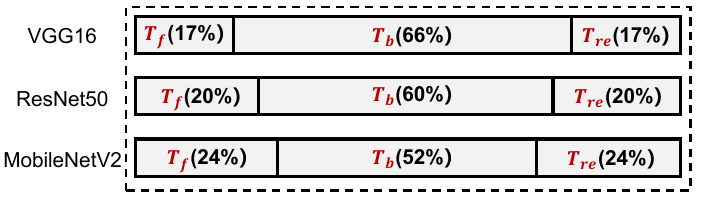}
  \caption{Latency breakdown of EcoTTA \cite{song2023ecotta}, a classic test-time adaptation scheme.}
  \label{fig:forward_backward_latency}
\end{figure}

\figref{fig:forward_backward_latency} show the latency breakdown of EcoTTA \cite{song2023ecotta}, a classic TTA method with three model architectures on NVIDIA Jetson NX, an off-the-shelf edge device.
The \textit{backward latency} $T_{b}$ takes up over $50\%$ of the overall delay, making it a \textit{bottleneck}.
This is because the forward pass only computes the activations, whereas the backward pass calculates the \textit{gradients} for both the \textit{activations} and \textit{parameters}, which doubles the \textit{computation} and demands \textit{extra memory access} to the activations from the forward pass \cite{li2020pytorch,moon2024new}.

\subsection{Problem Statement}
\label{sec:background:problem}
\label{sec:background:problem:basic}
\begin{figure}[t]
  \centering
  \includegraphics[width=.4\textwidth]{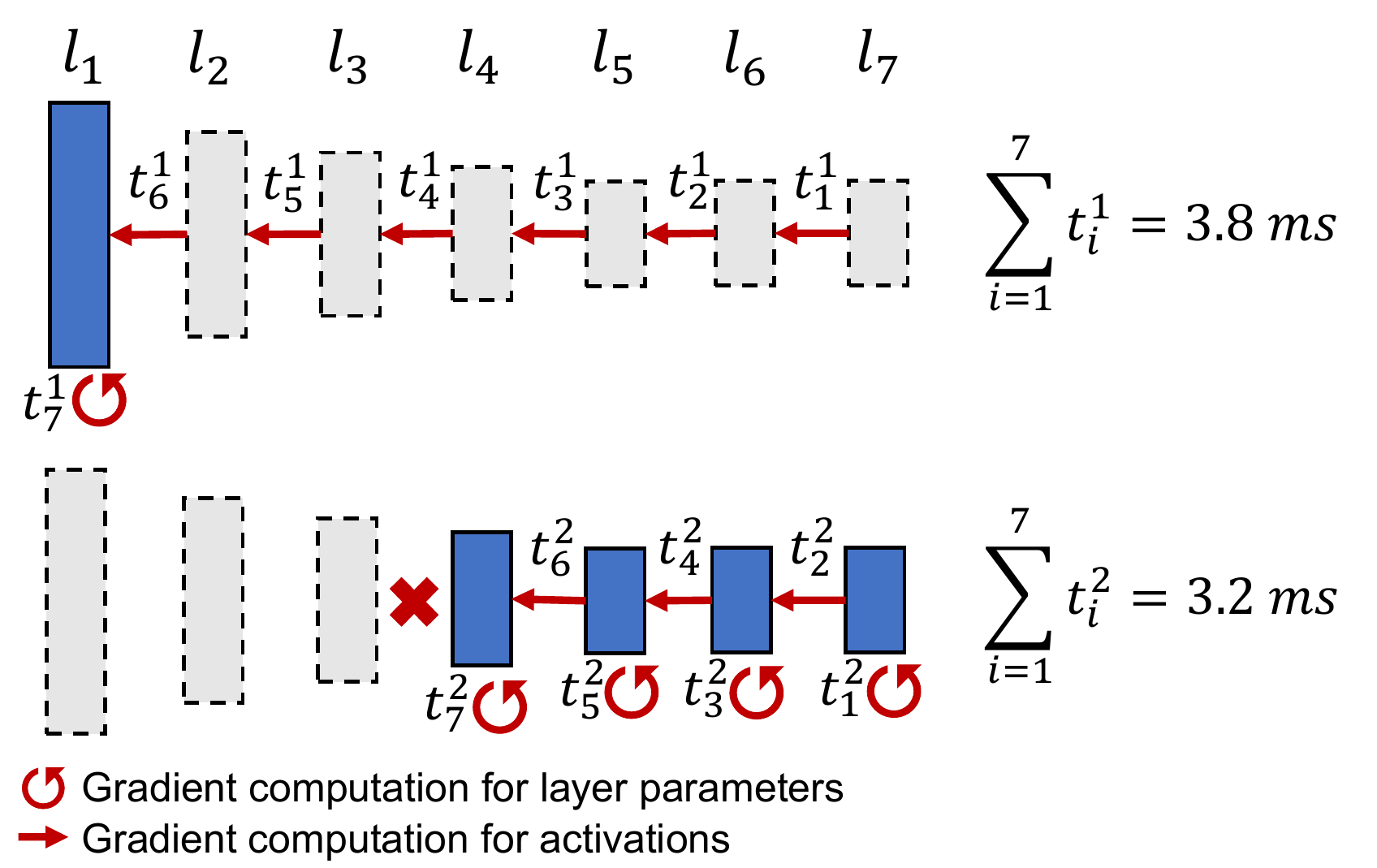}
  \caption{Example of sparse updating, where updating 1 or 4 layers results in almost the same latency.}
  \label{fig:layernum_latency}
\end{figure}

Our primary strategy to reduce adaptation latency is \textit{sparse updating} \cite{lee2019would, lin2022device, huang2023elastictrainer}, \ie selectively updating a subset of layers that are crucial for TTA.
It reduces the \textit{computation cost of gradients} and the \textit{memory accesses to retrieve intermediate activations}, thereby lowering latency in the backward pass.
However, we observe that \textit{updating fewer layers does not guarantee a proportional decrease in TTA latency on mobile devices}.
As illustrated in \figref{fig:layernum_latency}, the latency for updating the first layer is comparable to updating the last four layers. 
This is because updating the first layer still necessitates \textit{computing gradients} for all layers, leading to significant computation and memory access overhead.

Formally, we explore low-latency test-time adaptation to the unseen environment $e$ via sparse updating by formulating the following constrained optimization problem.
\begin{equation} \label{search_target}
    \max\text{\,\,}\vec{S}_e\cdot \vec{A}_e\,\,\,\,\,\,\,s.t.\,\,\,T_f+T_b^{'}( \vec{S}_e ) +T_{re}^{'}( \vec{S}_e ) \le \sigma \cdot T
\end{equation}
where $\vec{S}_{e}^{*}(N)$ is the optimal sparse updating strategy, which is a binary vector of length $N$ (the number of layers), and $\vec{A}_e$ is the layer importance vector of the same format as $\vec{S}_{e}^{*}(N)$.
$T_{b}^{'}$ and $T_{re}^{'}$ are the optimized backward and reforward latency under strategy $\vec{S}_{e}^{*}(N)$, respectively.
\lsc{$T$ is the overall latency without sparse updates, covering the time for a data batch to process from input to prediction results, including forward, backward, and reforward stages. }
$\sigma$ denotes the expected acceleration factor.
Note that $T_{f}$ is constant with diverse sparse update strategies. 
Sparse updating primarily reduces the backward latency, yet it also offers opportunities to decrease the reforward latency, as will be explained in \secref{sec:design:together}.
We formulate latency as a constraint with a tunable acceleration factor for easy configuration of the real-time requirements in diverse mobile applications with streaming data, \eg video processing~\cite{lu2019real,hu2017online} (image classification, object detection, scene understanding) of vehicle cameras or AR headsets~\cite{petrangeli2019dynamic,li2021lane,kong2023accumo}.

\begin{figure}[t]
  \centering
  \includegraphics[width=\linewidth]{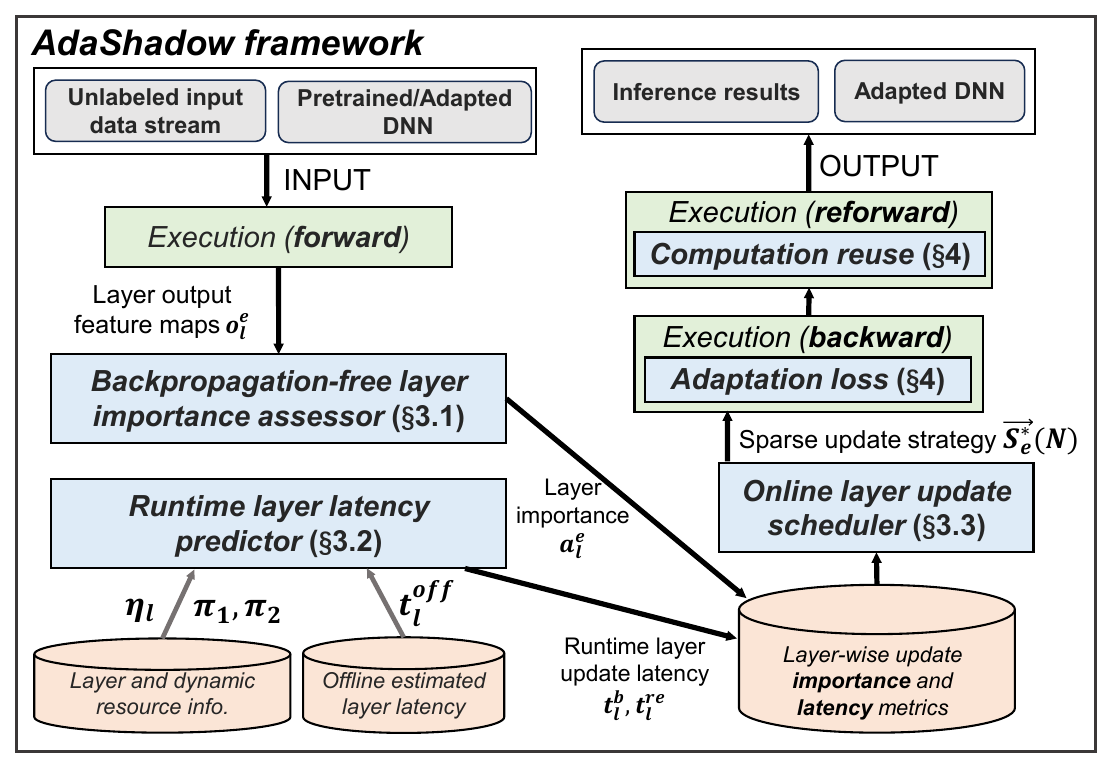}
  \caption{AdaShadow overview.}
  \label{fig:overview}
\end{figure}

\section{System Design}
\label{sec:design}

\noindent\textbf{System Overview}. 
\figref{fig:overview} shows the system architecture of \sysname. 
It consists of three functional modules:
\textit{(i)} the \textit{backpropagation-free layer importance assessor} resolves \textit{Challenge $\#1$} by evaluating layer importance via the \textit{divergence} of output feature embeddings across environments, utilizing \textit{unlabeled data} in the forward pass (\secref{sec:design:assessor}).
\textit{(ii)} the \textit{runtime layer latency predictor} overcomes \textit{Challenge $\#2$} by integrating mobile resource dynamics into the latency modeling for online calibration of offline estimates (\secref{sec:design:profiler}).
\textit{(iii)} the \textit{online layer update scheduler} addresses \textit{Challenge $\#3$} via a lightweight dynamic programming strategy to search for the optimal layer update strategy with high efficiency (\secref{sec:design:scheduler}).
Additionally, we present the implementation and additional optimization of the \sysname system workflow in \secref{sec:design:together}.


\subsection{Backpropagation-free Layer Importance Assessor}
\label{sec:design:assessor}
A prerequisite for sparse updating is to estimate the layer importance for on-device model adaptation.
This process should be \textit{latency-} and \textit{memory-efficient} to allow frequent revocation on low-resource mobile devices in non-stationary environments.
This subsection presents a backpropagation-free approach to significantly enhance the efficiency of prior schemes \cite{huang2023elastictrainer, lin2022device, qu2022p} for layer importance assessment.


\fakeparagraph{Limitations of Prior Arts} 
Existing studies \cite{huang2023elastictrainer, lin2022device} rely on \textit{gradients} to assess layer importance on adaptation accuracy, which faces two drawbacks. 
\textit{(i)} 
Accurate \textit{gradients} require \textit{labels}, which are absent in mobile-end TTA.
\textit{(ii)}
\lsc{
More critically, even with labels, the reliance on backpropagation to compute gradients for assessing layer importance results in substantial computational and memory overhead, as gradients must be propagated and stored for each layer. 
For instance, TTE \cite{lin2022device} and ElasticTrainer \cite{huang2023elastictrainer} incur latencies of up to $167.2s$ and $2.98s$, respectively, while consuming over 2,108 MB of memory for intermediate activations.
}
Such overheads are unacceptable for \textit{near-/real-time} DNN adaptation on mobile devices.

We propose to assess layer importance with \textit{unlabeled data} merely in the \textit{forward pass} for responsiveness by measuring the \textit{divergence} between the layers' \textit{output feature maps} in different environments.
The idea is inspired by the distribution alignment techniques \cite{schneider2020improving, mirza2022norm}, which quantifies drifts between the source and target data via the distributions of intermediate features in the forward stage. 
The \textit{rationale} is that when a DNN encounters domain shifts from a new environment, it will be less confident about its predictions, as reflected by a deviation in the layer's output distribution from that observed in historical environments.

\begin{table}[t]
\centering
\footnotesize
\caption{Overhead of layer importance assessors evaluated on Raspberry Pi.}
\renewcommand{\arraystretch}{1.1}
\scalebox{0.95}{
\begin{tabular}{|c|cc|cc|}
\hline
\textbf{}        & \multicolumn{2}{c|}{\textbf{Latency (s)}}                     & \multicolumn{2}{c|}{\textbf{Memory usage (MB)}}                \\ \hline
\textbf{Methods} & \multicolumn{1}{c|}{\textbf{ResNet50}} & \textbf{MobileNetV2} & \multicolumn{1}{c|}{\textbf{ResNet50}} & \textbf{MobileNetV2} \\ \hline
TTE \cite{lin2022device}              & \multicolumn{1}{c|}{167.24}            & 65.31                & \multicolumn{1}{c|}{2108}              & 695                  \\ \hline
ElasticTrainer \cite{huang2023elastictrainer}   & \multicolumn{1}{c|}{2.98}              & 1.29                 & \multicolumn{1}{c|}{2108}              & 695                  \\ \hline
\sysname         & \multicolumn{1}{c|}{\textbf{0.11}}              & \textbf{0.09}                 & \multicolumn{1}{c|}{\textbf{0.0086}}            & \textbf{0.0083}               \\ \hline
\end{tabular}
}
\label{tb:importance}
\end{table}

We cater this principle for TTA context as follows.
Given an input data batch $x_e$ from a new environment $e$, we assess the importance $a_l^e$ of layer $l$ in adapting to $x_e$ by measuring the Kullback-Leibler (KL) divergence between the embedding $E_{l}^{e}$ of layer $l$'s output feature $o_{l}^{e}$ on $x_e$, \ie  $E_{l}^{e} = g( o_{l}^{e} )$, and the embedding $E_{l}^{H}$ on the average output features on historical environments $H$.
Specifically, the layer importance $a_l^e$ is estimated as:
\begin{equation}\label{eq:a_l}
a_{l}^{e}=D_{KL}( g( o_{l}^{H} ) \lVert g( o_{l}^{e} ) )
\end{equation} 
where $D_{KL}(\cdot)$ denotes the KL divergence.
We justify our layer importance metric as follows.  
\begin{itemize}
    \item 
    The \textit{KL divergence} effectively quantifies the data distribution shift in mobile environment $e$ from the history $H$ perceived at layer $l$.
    A larger divergence indicates a greater necessity to update layer $l$.
    Since we target at adaptation on data streams, \eg live videos, the domain shifts between adjacent batches tend to be mild.
    \lsc{
    Hence KL divergence can effectively handle mild domain shifts, it is unnecessary to adopt computation-intensive metrics, \eg the Wasserstein distance, which are intended for distributions with low overlaps \cite{shen2018wasserstein}.
    }    
    
    \item 
    Rather than the raw output features, we measure the divergence of their \textit{embedding} via an embedding function $g(\cdot)$.
    It normalizes output features to the same dimensions for fair importance evaluation across layers.
    More critically, it allows robust estimation with small batches, which is essential for non-stationary environments, as will be discussed next.
\end{itemize}

Furthermore, we implement the layer importance metric with two optimizations for mobile environments.
\begin{itemize}
    \item 
    For efficient distribution estimation with \textit{small-batch} mobile data, we devise $g(\cdot)$ with only first and second-order moments.
    \lsc{Compared to existing methods \cite{zellinger2017central,chen2020homm} that minimize differences in higher-order sample moments, it can effectively extracts sufficient distribution information from small batch data.}
    Concretely, we compute the channel-wise means $\mu (o_{l,c}^{e} )$ = $\frac{1}{N}\sum_{c\in \mathbb{C}}{o_{l}^{e}}$, and the channel-wise variances $\sigma ^2( o_{l,c}^{e} )$ = $\frac{1}{N}\sum_{c\in \mathbb{C}}{( o_{l}^{e}-\mu ( o_{l,c}^{e} ) ) ^2}$ from layer $l$'s output features $o_{l,c}^{e}$.
    The embedding $E_{l}^{e}$ is then represented as $[ \mu ( o_{l,1}^{e} ) , \sigma ^2( o_{l,1}^{e} ), \ldots, \mu ( o_{l,N_c}^{e} ), \sigma ^2( o_{l,N_c}^{e})]$.
    \item
    To track the \textit{non-stationary} mobile environments, we gradually update the historical embedding $E_l^H = g( o_{l}^{H} )$ by integrating the new environment $e$ into $H$ via a moving average strategy: $g( o_{l}^{H} ) =\alpha \cdot g( o_{l}^{e} ) +( 1-\alpha) \cdot g( o_{l}^{H} )$, where $\alpha \in [ 0,1 ]$ is a hyperparameter controlling the incorporating rate of new environment $e$.
\end{itemize}

\begin{figure}[t]
  \centering
  \includegraphics[width=0.45\textwidth]{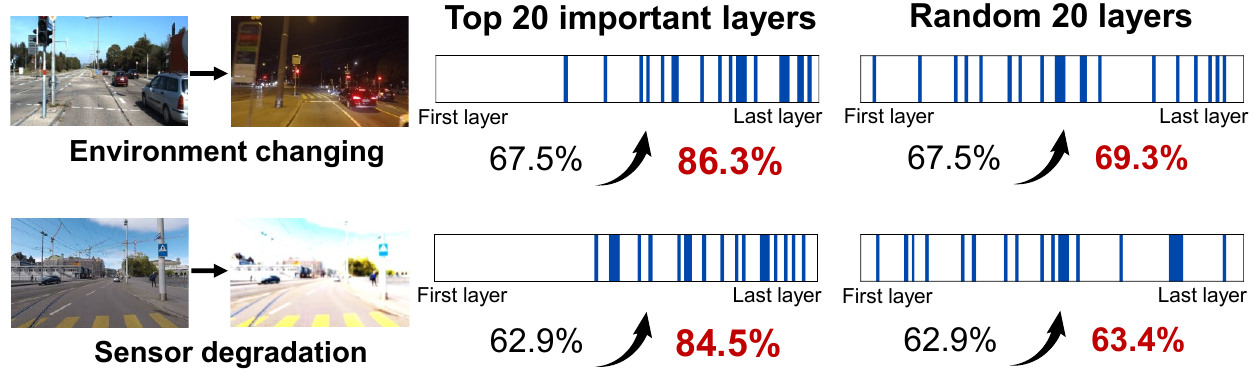}
  \caption{Accuracy gains when updating top 20 important layers or 20 random layers (blue bars).}
  \label{fig:assessor_accuracy}
\end{figure}

\lsc{
The layer importance assessor reuses existing embeddings, eliminating the need for additional forward propagation. 
This limits the computational cost to statistic extraction and KL divergence calculation. 
For example, in ResNet50, our importance profiler incurs only 0.2 GFLOPs overhead for these computations, significantly lower than the 4.1 GFLOPs saved by halving backpropagation when \(\sigma = 0.5\). 
Consequently, the theoretical latency can be optimized to as low as 4.8\% of the original (0.2 GFLOPs) or up to 104.8\% of the original (4.3 GFLOPs) when \(\sigma = 1\).
}
As shown in Tab. \ref{tb:importance}, our layer importance assessor reduces evaluation latency by 95\% compared to gradient-based methods like TTE \cite{lin2022device} and ElasticTrainer \cite{huang2023elastictrainer}. 
It also has low memory requirements, using less than $8.6$ KB to store historical embeddings for ResNet50. 
Importantly, this efficiency does not compromise accuracy; our method accurately identifies the importance of individual layers for TTA, as illustrated by the blue bars in Fig. \ref{fig:assessor_accuracy}, which represent the layers selected for updating.

\subsection{Runtime Layer Latency Predictor}
\label{sec:design:profiler}
Precise layer update latency estimation is another key input to determine the optimal sparse updating strategy.
As we target at TTA on mobile devices with limited and dynamic resources, \textit{offline} predicting schemes \cite{zhang2021nn, huang2023elastictrainer} incur large estimation errors.
This subsection introduces an \textit{accurate} and \textit{lightweight} approach to model and calibrate the backward\footnote{Although we optimize both the latency during backward and reforward (see \equref{search_target}), we mainly model and calibrate the estimation on backward latency for two reasons. 
\textit{(i)} The backward pass is the latency bottleneck (see \secref{sec:background:latency:bottleneck}).
\textit{(ii)} In practices, the reforward latency can be easily hidden via pipelining adaptation and inference (see \secref{sec:exp:case}).} latency of individual layers to \textit{runtime} resource dynamics.

\begin{figure}[t]
  \centering
  \includegraphics[width=0.35\textwidth]{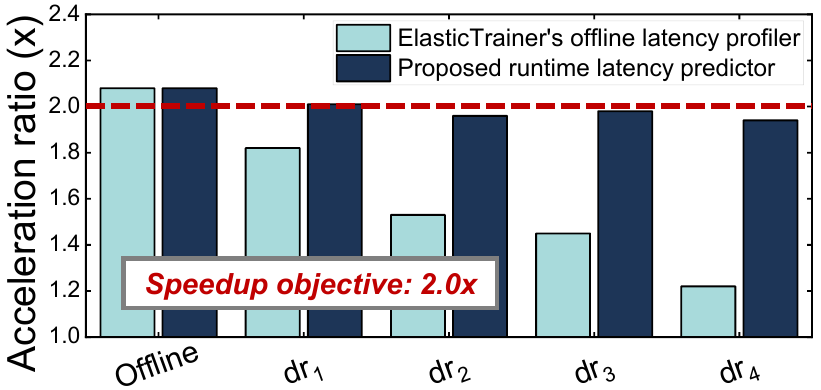}
  \caption{Comparison of offline/runtime latency predicting with dynamic resource availability.}
  \label{fig:Dynamic_hardware_condition}
\end{figure}

We empirically demonstrate the necessity of estimating the training latency at runtime.
Specifically, we first apply ElasticTrainer~\cite{huang2023elastictrainer}, the state-of-the-art sparse training scheme that predicts layer latency offline, to speed up the adaptation of a ResNet50 and NVIDIA Jetson NX by an acceleration ratio of about $2\times$ under \textit{abundant resources} (denoted as \textit{offline} in \figref{fig:Dynamic_hardware_condition}).
Then we simulate four dynamic resource conditions $dr_1 \sim dr_4$ commonly seen in mobile systems \cite{she2023accurate}: $dr_1$: increase the mobile CPU/GPU temperature to 60 \textdegree C; $dr_2$: add three competing compute-intensive processes; $dr_3$: occupy the cache to tune the cache hit rate to be $30\%$; $dr_4$: combine all factors from $dr_1 \sim dr_3$.
When running ElasticTrainer~\cite{huang2023elastictrainer} in these four mobile system conditions using configurations obtained in the offline case, we observe mild to drastic decrease in acceleration ratios (see \figref{fig:Dynamic_hardware_condition}).
The experiment shows that the \textit{offline layer latency} estimates can be significantly \textit{inaccurate} due to dynamic resource availability. 
\cite{she2023accurate} incorporates dynamic resources into latency predictions using a Graph Neural Network (GNN), however, incurring substantial overhead.
Furthermore, latency profiling is ineffective without actual execution, as noted in \cite{chu2023nnperf}.
In contrast, our proposed method delivers higher acceleration ratios by adapting to these runtime conditions.

\textbf{Key Idea}.
We observe that the execution latency of a DL unit mainly hinges on on-chip kernel computation and off-chip memory access. 
Each unit's kernel execution strategy and memory allocation are static, whereas kernel utilization, temperature, and cache-hit-rate are dynamic and measurable at runtime. 
Moreover, this unit-based approach is versatile across different DL models. For example, basic units in ResNet include Conv2d, BatchNorm, and Linear, while in Transformer, they consist of projectors Q, K, V, LayerNorm, and the feed-forward network (FFN).

\fakeparagraph{Methods}
In offline predicting with \textit{abundant}, \textit{static} resources, the latency $t_{l}^{off}$ of layer $l$ can be estimated as $t_{l}^{off}$ = $t_{l}^{c_{off}}$ + $t_{l}^{m_{off}}$, where $t_{l}^{c_{off}}$ and $t_{l}^{m_{off}}$ are corresponding latency for computation and memory access profiled offline, respectively.
We account for \textit{limited}, \textit{dynamic} runtime resources on device by modeling the layer latency $t_{l}$ = $\pi_1\cdot t_{l}^{c_{off}}$ + $\pi_2\cdot t_{l}^{m_{off}}$, where $\pi_1>1$ and $\pi_2>1$ are the computation and memory expansion coefficients.
As measuring $t_{l}^{c_{off}}$ and $t_{l}^{m_{off}}$ separately is challenging, we correlate $t_{l}$ to $t_{l}^{off}$ as follows.
\begin{equation}\label{eq:t_l}
    t_{l} = (\pi_1\cdot \frac{\eta_l}{\eta_l+1}+\pi_2\cdot \frac{1}{\eta_l+1}) \cdot t_{l}^{off}
\end{equation}  
Note that $t_{l}^{off}$ is measurable offline.
\lsc{
To simplify Eq. \ref{eq:t_l}, we define $\eta_l = \frac{t_{l}^{c_{off}}}{t_{l}^{m_{off}}}$, a layer-dependent metric that remains constant regardless of runtime resource conditions. 
This metric captures the effects of the computation expansion coefficient $\pi_1$ and the memory expansion coefficient $\pi_2$ on overall unit latency. 
By using distinct coefficients for computation and memory, we can effectively profile their impacts on runtime performance.
}
The two coefficients $\pi_1$ and $\pi_2$ reflect resource limitations on computation and memory latency, applicable across layers and measurable during runtime. We'll detail how to assess these three hyperparameters next.

\begin{figure}[t]
  \centering
  \includegraphics[width=0.47\textwidth]{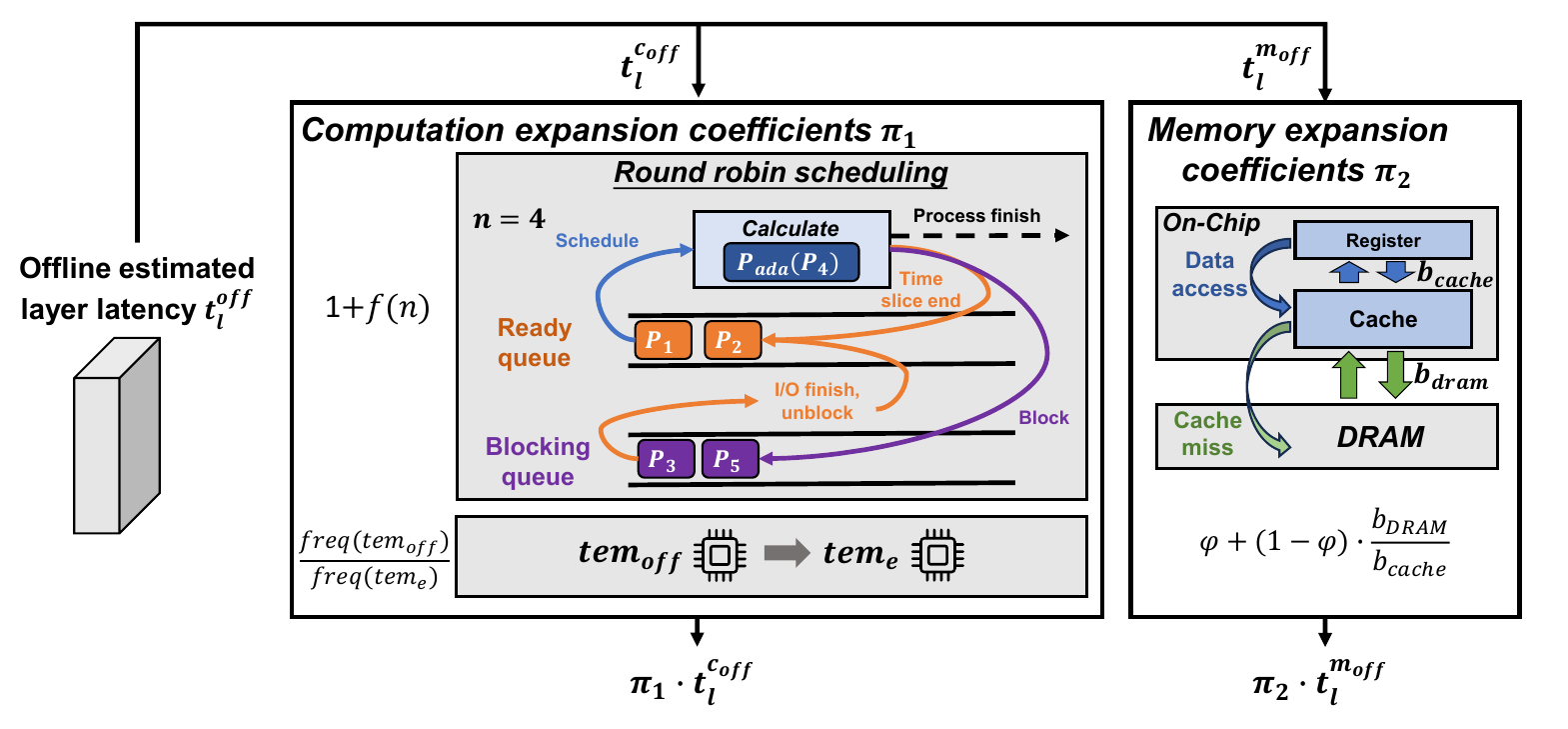}
  \caption{Layer update latency in offline and online.}
  \label{fig:layer_latency_compose}
\end{figure}

\noindent$\bullet$ \textit{Profiling $\pi_1$}.
    The computation expansion coefficient $\pi_1$ depicts two runtime factors that affect computation latency: \textit{(i)} number $n$ of competing processes, and \textit{(ii)} core temperature $tem^{on}$ of mobile CPU/GPU.
    \begin{itemize}
        \item 
        [-] The OS kernel of the mobile CPU \cite{balharith2019round, singh2010optimized} and GPU \cite{verner2012scheduling, pattnaik2016scheduling} often employs round robin scheduling (see \figref{fig:layer_latency_compose}). 
        Hence an increase in process number leads to a linear growth in process waiting time. 
        \item
        [-] Dynamic Voltage Frequency Scaling (DVFS) \cite{kodukula2023squint, lin2023workload,herbert2007analysis} is often activated in the mobile CPU/GPU to avoid overheating by reducing the clock frequency. 
        This also increases the computation time.
    \end{itemize}
    Accordingly, we set $\pi_1=\frac{freq(tem^{off})}{freq(tem^{on})}\cdot [ 1+f(n)]$, where $f(\cdot)$ is a linear function determined offline depicting the process switching overhead.
    The dynamic frequency $freq(\cdot)$ accounts for the DVFS.

\noindent$\bullet$ \textit{Profiling $\pi_2$}.
    We relate the memory expansion coefficient $\pi_2$ to the cache-hit-rate $\varphi$, which is directly measurable at runtime.
    This is because sharing the limited cache among processes increases the memory access latency.
    Also, in case of cache miss, there is an extra latency for data movement, which is determined by the ratio between the bus bandwidth $b_{DRAM}$ and the cache bandwidth $b_{cache}$ (see \figref{fig:layer_latency_compose}).
    Hence, we set $\pi _2=\varphi + (1-\varphi)\cdot\frac{b_{DRAM}}{b_{cache}}$.

\noindent$\bullet$  \textit{Profiling $\eta_l$}.
    Although it is challenging to measure $t_{l}^{c_{off}}$ and $t_{l}^{m_{off}}$ separately, their ratio $\eta_l=\frac{t_{l}^{c_{off}}}{t_{l}^{m_{off}}}$ can be profiled offline.
    Specifically, we transform $\eta_l=\frac{c^l\cdot\delta_c}{m^l\cdot\delta_m}$, where $c^l$, $m^l$, $\delta_c$, and $\delta_m$ denote amount of MAC, amount of memory accesses, unit MAC time, and unit memory access time of layer $l$, respectively.
    $c^l$ and $m^l$ can be directly derived according to the type and hyperparameters of layer $l.$
    $\delta_c$ is obtained from the processor's FLOPS $F$: $\delta _c=\frac{1}{F}$, while $\delta_m$ is obtained from the cache bandwidth $b_{cache}$: $\delta _m=\frac{1}{b_{cache}}$.    
    We also include non-parameterized layers, \eg activation layers, in $t_{l}^{c_{off}}$ as they also involve in gradient computation.
\lsc{
The cost of latency prediction only involves sensing dynamic resource states and calculating execution latency using Eq. \label{eq:t_l}, resulting in negligible computational and memory overhead.
}



\subsection{Online Layer Update Scheduler}
\label{sec:design:scheduler}
This subsection presents a dynamic programming (DP) based lightweight online scheduler that efficiently decides the best layer update strategy $\vec{S}_{e}^{*}(N)$ for \equref{search_target}, given the layer-wise metrics of a DNN.
Picking the optimal solution to \equref{search_target}, an NP-hard nonlinear integer programming problem, can be time-consuming\cite{lin2022device}; 
We formulate specific subproblems, recursion, and space for three-stage TTA.


Consider a DNN with \(N\) layers. 
Let the importance and update latency of layer \(l\) (counting from the last layer) be \(a_l^e\) and \(t_l^e\) in environment \(e\). 
A brute-force search for the optimal update strategy \(\vec{S}_{e}^{*}(N)\) has a complexity of \(O(2^N)\), which is impractical for online scheduling. 
To address this, we propose a dynamic programming (DP) formulation to solve \equref{search_target}. 
Specifically, we aim to find the optimal layer selection \(\vec{S}_{e}^{*}(N)\) that maximizes cumulative importance within the latency budget \(T_{bgt} = \sigma \cdot T - T_f\). 
Let \(P[l][t]\) represent the maximum cumulative layer importance from the last layer to layer \(l\) for a given latency budget \(t\). 
The binary vector \(\vec{S}_e[l][t]\) corresponds to one solution for \(P[l][t]\), while \(\vec{S}_{e}^{*}[l]\) denotes the optimal solution. 
We formulate the DP at the granularity of \(t\) and \(l\).

\begin{itemize}
    \item \textit{Discrete time}:
    We discretize time into $N_T$ units, \ie at a resolution of $T_{bgt}/N_T$.
    We empirically set $N_T=500$ which effectively balances search cost and accuracy.
    \item \textit{Layer-wise}: 
    It is natural to decompose sub-problems layer-wise to support diverse DNN architectures.
    It also aligns with the layer-wise DNN execution scheme on mobile devices with limited resources \cite{zhang2021nn}. 
    Finer-grained decomposition \eg tensor-wise as \cite{huang2023elastictrainer} can be inefficient ($214\cdot N_T$ sub-problems in tensor-wise vs. $107\cdot N_T$ sub-problems in layer-wise for ResNet50).
    
\end{itemize}

\begin{figure}[t]
  \centering
  \includegraphics[width=0.45\textwidth]{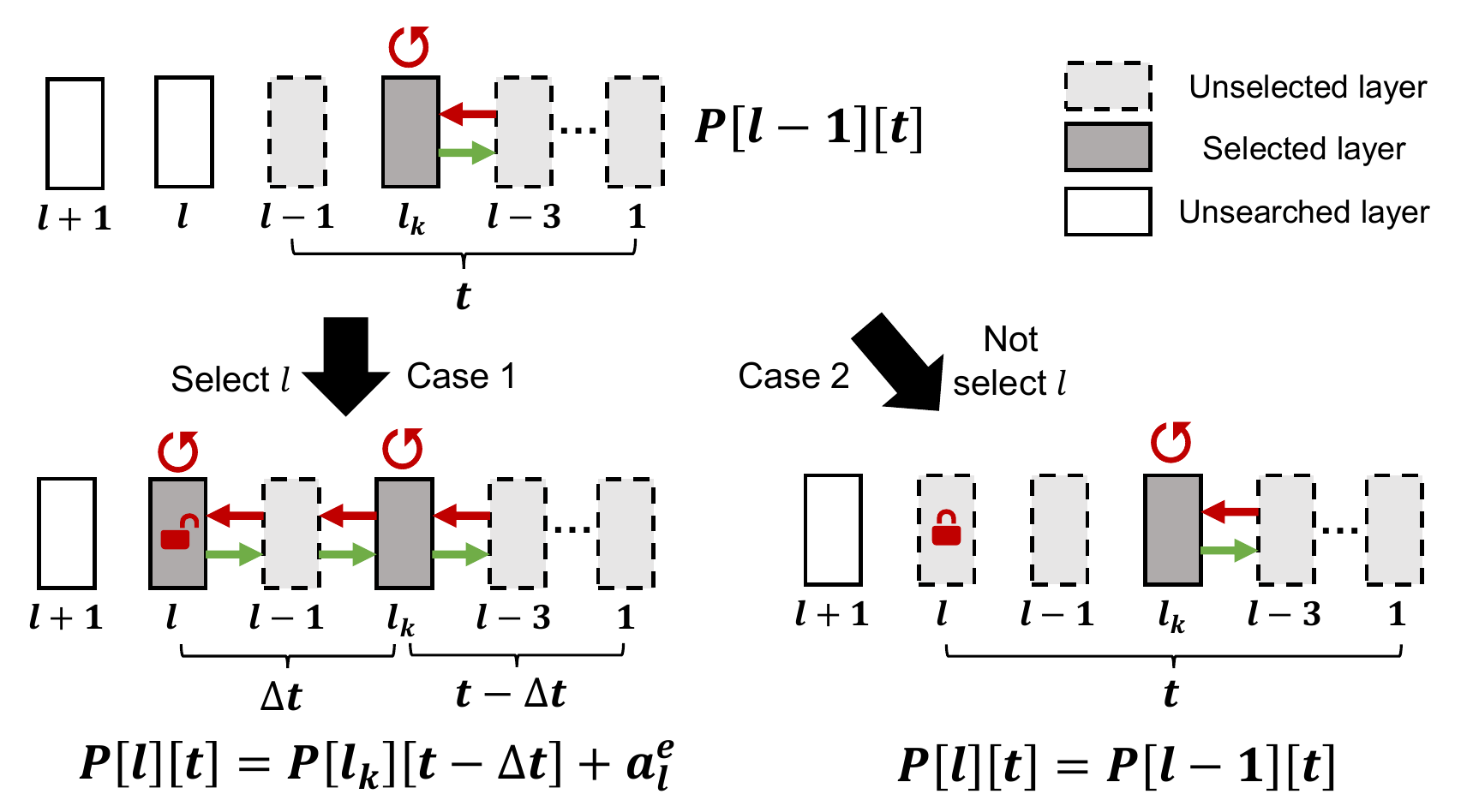}
  \caption{Illustration of recursion in two different cases.}
  \label{fig:dp_algorithm}
\end{figure}

Accordingly, $0 \le l \le N$ and $0 \le t \le N_T$, and the DP formulation would reduce the time complexity to $O(N^2 \cdot N_T)$ considering the time complexity $O(N)$ of solving each sub-problem (details below). 

The base cases are trivial. 
We set $P[0][t]$ for all $0 \le t \le N_T$ as $0$.
This is because the maximum cumulative layer importance is $0$ if no layer is selected for updating.
The recursion for $P[l][t]$ when $l > 0$ is more subtle.

\textit{i) \textbf{Case 1}}:
    If layer $l$ is selected, then the maximum cumulative layer importance becomes $P[l_k] [t-\varDelta t] +a_{l}^{e}$, where layer $l_k$ is the last layer selected in $P[ l-1][ t]$, \ie the closest selection to layer $l$.
    This is because updating one more layer decreases the latency budget when backpropagating till layer $l-1$ by $\varDelta t$ (details below) and increases importance by $a_{l}^{e}$.

\textit{ii) \textbf{Case 2}}:
    If layer $l$ is not selected, the maximum cumulative layer importance remains $P[l-1][t]$.

As shown in \figref{fig:dp_algorithm}, we clarify the two cases assuming $l_k = l-2$.
Since we search for the maximum, the final recursion becomes: 
$P[ l,t ] =\max \{ P[ l-1,t ] ,\,\,P[ l_k,t-\varDelta t ] +a_{l}^{e} \}$.
Note that the extra latency $\varDelta t$ induced by selecting layer $l$ in \textit{Case 1} is not simply its layer update latency $t_l^e$.
From \figref{fig:dp_algorithm}, $\varDelta t=t_{l}^{\delta w}+\sum_{m={l_k}}^{l}{t_m^{\delta x}}+\sum_{m={l_k+1}}^{l}{t_{m}^{re}}$, where $t_{l}^{\delta w}$ is the gradient calculation time of weight $w_{l}$, $\sum_{m={l_k}}^{l-1}{t_m^{\delta x}}$ is the gradient calculation time of activations between layer $l-1$ and layer $l_k$, and $\sum_{m={l_k+1}}^{l}{t_{m}^{re}}$ is the reforward time between layer $l$ and layer $l_k+1$. 
Noting that $t_{l}^{b}=t_{l}^{\delta x}+t_{l}^{\delta w}$, we utilize the amount of MAC of $t_{l}^{\delta x}$ and $t_{l}^{\delta w}$ to infer their proportion of latency in $t_{l}^{b}$.
As $l_k$ is unknown in advance, we gradually decrease $l_k$ from $l-1$ to $0$ to explore all possible $l_k$ to find the $l_k$ that maximizes $P[ l_k][t-\varDelta t ]$ with a time complexity of $O(N)$.

Given the above formulations, we can explore all $P[l][t]$ recursively till $P[N][T_{bgt}]$, and then we can search the optimal layer update strategy $\vec{S}_{e}^{*}[N]$.
To further improve the efficiency, we refine the search space by removing two types of invalid sub-problems. 

\begin{itemize}
    \item For $T_{bgt}$, we iteratively exclude sub-problem $P[l,t]$ if $t$ exceeds the latency budget $T_{bgt}$.We can also discard all sub-problems generated from an invalid $P[l,t]$.
    \item For $t$ in sub-problem $P[l,t]$, if the gradient computation time $\sum_{m=1}^l{t_{\delta x}^{m}}$ exceeds $t$, the total latency of $P[l,t]$ will also exceed $t$ even if no layers are selected for updating.Hence, we can discard such $P[l,t]$ and all sub-problems it generates.
\end{itemize}

To eliminate redundant calculations, we identify overlapping solutions by formulating the latency of both backward and reforward into the same recursive sub-problem. 
This is because, as noted in \secref{sec:background:problem:basic}, the \textit{position of the first update layer} impacts the latency of both backward and reforward simultaneously.



\section{Implementation}
\label{sec:design:together}

This subsection presents how the layer importance assessor (\secref{sec:design:assessor}), layer latency predictor (\secref{sec:design:profiler}), and layer update scheduler (\secref{sec:design:scheduler}) cooperates in the \textit{forward-backward-reforward} pipeline (see \figref{fig:overview}) as well as  our \textit{additional optimizations} to improve the accuracy and efficiency of TTA on mobiles.

At initialization, DNNs are pre-trained in the cloud using source datasets and then deployed to mobile devices.
Upon receiving a batch of test samples $x_e$ from a new environment $e$, \sysname adapts the DNN as follows.
\begin{itemize}
    \item \textit{Forward phase}.
    \sysname employs the layer importance assessor and the latency predictor to derive the layer importance $a_l^e$ and the backward and reforward latency $t_l^b$ and $t_{l}^{re}$ for each layer of the DNN.
    Meanwhile, it prepares each layer's output feature maps $o_{l}^{e}$ and computation graph for the backward phase.
    \item \textit{Backward phase}. 
    \sysname calls the layer update scheduler to search for the optimal strategy $\vec{S}_{e}^{*}(N)$, based on $a_l^e$, $t_l^b$, and $t_l^{re}$ obtained in the forward stage.
     \sysname prunes computation graph nodes of layers not in $\vec{S}_{e}^{*}(N)$ to skip gradient calculation at the compiler level, to realize the selective propagation process.
    The training loss will be discussed later.
    \item \textit{Reforward phase}. 
    \sysname infers on $x_e$ using the updated model from the backward stage.
    \sysname partially reuses computation results from the forward stage to accelerate the inference, as described next.
\end{itemize}

We propose two more optimizations during backward and reforward to boost the performance.

\noindent$\bullet$ \textit{Adaptation loss for small batches}.
    Existing TTA methods \cite{wang2021tent, niu2023towards} mainly use the \textit{entropy} of the \textit{final layer} output as the unsupervised adaptation loss, often resulting in low accuracy with small batches \cite{niu2023towards}. 
    In \sysname, we refine the loss $L$ by summing the KL-divergence between the new environment and the history from \textit{all layers}, \ie $L=\sum_{l=1}^N{D_{KL} ( E_{l}^{H}\lVert E_{l}^{e})}.$ 
    Evaluations show that the refined loss is more robust to small batch sizes.

\begin{figure}[t]
  \centering
  \includegraphics[width=0.4\textwidth]{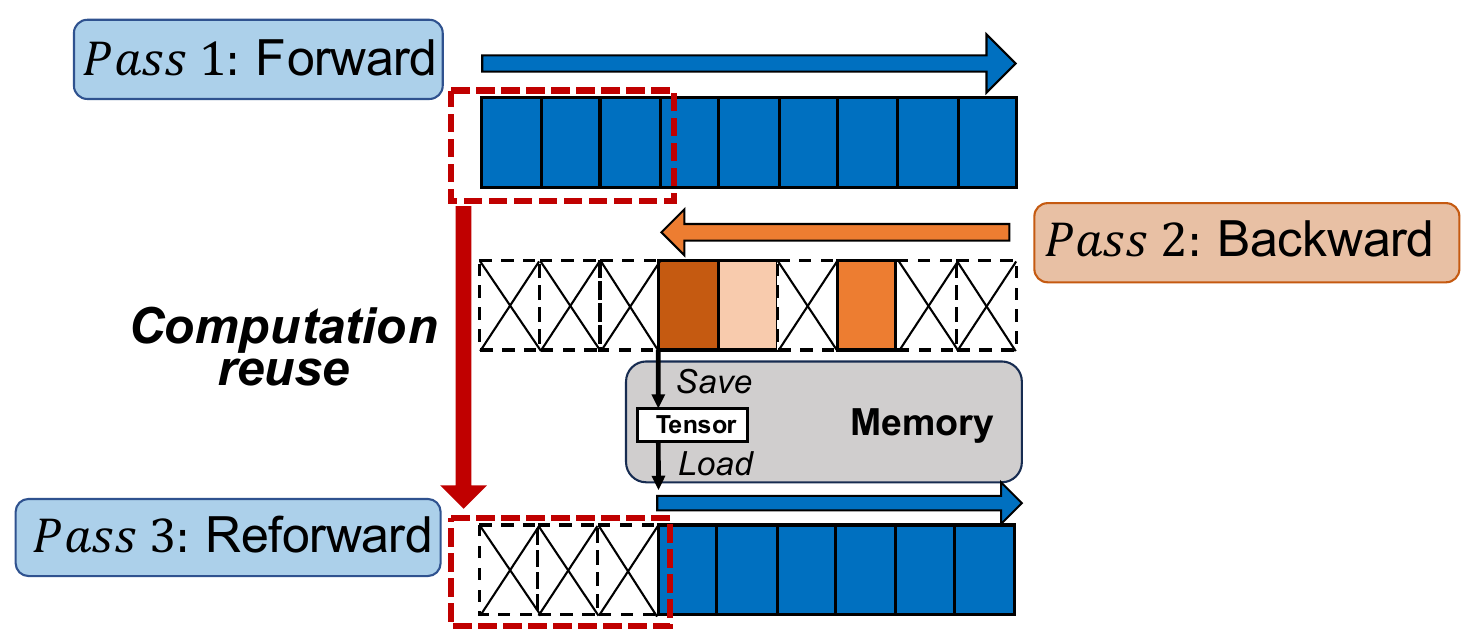}
  \caption{Illustration of memory I/O-aware computation reuse implementation.}
  \label{fig:tensor_save_load}
\end{figure}

\noindent$\bullet$ \textit{Memory I/O-aware computation reuse between forward and reforward}.
    Since both forward and reforward computations are performed on the same batch, we can partially reuse computations from the forward stage to reduce overall latency, as illustrated in Fig. \ref{fig:tensor_save_load}. 
    We track memory addresses for input tensors during forward. 
    After defining the layer update strategy, we detach the input activation $x_n$ of the first updated layer $n$ in $\vec{S}_{e}^{*}(N)$ from the gradient computation graph and retain it in memory, while releasing input tensors $x_i$ for other layers. 
    Before reforward begins, we load $x_n$ from memory and reuse computation results for layers not updated during the backward pass, skipping layers before the first updated layer $n$ during reforward.

\section{Evaluation}
\label{sec:exp}
\subsection{Experiment Setup}

\fakeparagraph{Implementation} 
We prototype \sysname on three mobile devices.
$D_1$: NVIDIA Jetson NX (Cortex-A57 CPU and 256 CUDA cores GPU); $D_2$: NVIDIA Jetson Nano (Cortex-A57 CPU and 128 CUDA core GPU); $D_3$: Raspberry Pi 4B (Cortex-A72 CPU). 
We installed Jetpack 4.6.4 OS based on Ubuntu 18.04 on $D_1$ and $D_2$, and the 64-bit Raspbian 11 OS on $D_3$.
Before DNN deployment, \sysname measures the multi-process scheduling latency function \(f( n )\) and the offline layer execution latency via one pass of backpropagation using PyTorch profiler plugin, and detects frequency-temperature function \(freq( tem )\) through the device DVFS strategy.
During test-time adaptation, \sysname uses the SGD optimizer with a learning rate $lr$ of 5$e$-3. 


\fakeparagraph{Tasks, Datasets, and Models}
We test three mobile sensing tasks with diverse continual domain shifts.
\begin{itemize}
    \item \textbf{NICO++ \cite{zhang2023nico++}}. 
    It covers 6 types of \textit{environmental} shifts, including outdoor, autumn, dim, grass, rock, and water.
    The model is pre-trained on 'outdoor' for adaptation to other five environments.
    NICO++ are commonly used datasets for evaluating domain adaptation \cite{zhang2023flatness, gui2024joint}.
    
    \item \textbf{CIFAR10-C and CIFAR100-C \cite{hendrycks2019benchmarking}}.
    CIFAR10-C and CIFAR100-C has been widely used for evaluation adaptation performance in mobile systems~\cite{deng2022geryon,ghosh2023energy,ding2022towards}. 
    They simulate 15 types of \textit{corruption} shifts, induced by lens shakes, low-light conditions, and sensor degradation such as Gaussian noise, motion blur, pixelation, \etc
    We experiment with two dataset combinations: [CIFAR10, CIFAR10-C] and [CIFAR100, CIFAR100-C], where the first is for pre-training and the second for adaptation.
    
    \item \textbf{Self-collected mobile videos}.
    It contains 10 natural, continual, potentially \textit{mixed} shifts collected in open mobile environments.
    Specifically, we built an Ackermann vehicle platform equipped with Jetson NX development board, STM32 control board, visual sensors, \etc to move on campus and collect 10 hours of videos.
    We manually sample video clips that contain transitions of shifts, \eg indoor-wild, sunny-rainy-cloudy, daytime-night, and different view angles, \etc
    \item \textbf{Self-collected layer adaptation latency records}.
    We build a dataset of 15,000 records capturing various real-world system factors and layer adaptation latency.
    
\end{itemize}

We adopt three typical model architectures, \ie vision transformer (ViT-B/16)~\cite{dosovitskiy2020image}, VGG16 (without BN layer)~\cite{simonyan2014very}, MobileNetV2 (with BN)~\cite{sandler2018mobilenetv2}, and ResNet50 (with BN)~\cite{he2016deep}.

\fakeparagraph{Baselines} 
\label{Baselines}
Generic and supervised TTA set strict accuracy lines, while efficient TTA set latency benchmarks.
\begin{itemize}        
    \item \textbf{Generic TTA (no target label)}.
    They aim at accurate TTA without explicit efficiency optimization.
    \begin{itemize}
        \item[-] \textbf{Tent} \cite{wang2021tent}: the first TTA that updates the \textit{BN} layers via entropy minimization.
        \item[-] \textbf{CoTTA} \cite{wang2022continual}: another representative TTA strategy via \textit{knowledge distillation} for anti-forgetting.
    \end{itemize}
    \item \textbf{Efficient TTA (no target label)}.
    Besides accuracy, this category optimizes efficiency metrics of TTA.
    \begin{itemize}
        \item[-] \textbf{EATA} \cite{niu2022efficient}: \textit{data-efficient} TTA by updating BN layers on selective samples.
        \item[-] \textbf{EcoTTA}~\cite{song2023ecotta}: typical \textit{memory-efficient} TTA that reduces the memory footprint to update the BN layers with lightweight adapters.
        \item[-] \textbf{MECTA}~\cite{hong2023mecta}: \textit{memory-efficient} TTA that selectively updates the proposed \textit{adaptive BN} layers.
        \item[-] \textbf{BFTT3D}~\cite{wang2024backpropagation}: \textit{backpropagation-free} TTA via comparing source and target features. 
        \item[-] \textbf{LAME}~\cite{boudiaf2022parameter}: a \textit{latency-efficient} TTA that adapts the model’s output rather than parameters using manifold-regularized likelihood loss.
        \item[-] \textbf{OFTTA}~\cite{wang2024optimization}: a \textit{latency-efficient} TTA that selectively adjust the proposed \textit{exponential decay BN} layers.

    \end{itemize}
    \item \textbf{Efficient Training (supervised by target labels)}.
    This category focuses on efficient training on low-resource mobile and embedded devices.
    They provide unsupervised loss (\ie entropy) at test time.
    \begin{itemize}
        \item[-] \textbf{Melon} \cite{wang2022melon}: a \textit{memory-efficient} on-device training scheme that uses recomputation and micro-batch to fit into memory-scarce mobile devices.
        \item[-] \textbf{TTE}~\cite{lin2022device}: a \textit{memory-efficient} on-device training scheme that selectively updates tensors via \textit{offline} profiling.
        \item[-] \textbf{ElasticTrainer}~\cite{huang2023elastictrainer}: a \textit{latency-elastic} on-device training method that uses \textit{runtime} profiling to identify important tensors and \textit{offline} latency profiling, with the importance metric based on \textit{backpropagation}.
    \end{itemize}
\end{itemize}

\fakeparagraph{Process Schedule Modes} 
EATA(P) and EcoTTA(P) refer to the EATA and EcoTTA baselines implemented in parallel adaptation/inference mode. 
In contrast, other baselines operate in sequential adaptation/inference mode, applying the newly adapted model to subsequent sample inference for accuracy testing, as discussed in $\S$ \ref{sec:background:latency:bottleneck}.

\begin{table*}[t]
\centering
\footnotesize
\caption{Test accuracy (\%), adaptation throughput (fps), and latency (ms) with continual environmental shifts.}
\renewcommand{\arraystretch}{1.2}
\scalebox{1}{
\begin{tabular}{ccccccccccccccc}
\hline
\multicolumn{2}{c}{\textbf{Time}} & \multicolumn{10}{c}{\textbf{Continual data shift over time —\textgreater{}}} &  &  &  \\ \cline{1-12}
\multicolumn{2}{c}{\textbf{Round}} & \multicolumn{5}{c}{\textbf{1}} & \multicolumn{5}{c}{\textbf{2}} &  &  &  \\ \cline{1-12}
\textbf{Dataset} & \textbf{Method} & \textbf{Autumn} & \textbf{Dim} & \textbf{Grass} & \textbf{Rock} & \textbf{Water} & \textbf{Autumn} & \textbf{Dim} & \textbf{Grass} & \textbf{Rock} & \textbf{Water} & \multirow{-3}{*}{\begin{tabular}[c]{@{}c@{}} \textbf{Mean} \\ \textbf{ACC.(\%)} \end{tabular}} & \multirow{-3}{*}{\begin{tabular}[c]{@{}c@{}} \textbf{Th.} \\ \textbf{(fps)} \end{tabular}} & \multirow{-3}{*}{\begin{tabular}[c]{@{}c@{}} \textbf{Latency} \\ \textbf{(ms)} \end{tabular}} \\ \hline
 & TENT~\cite{wang2021tent} & 84.2 & 81.8 & 86.6 & 86.2 & 86.5 & 89.9 & 89.1 & 91.3 & 90.9 & 91.7 & 87.8 & 30.1 & 33.2 \\
 & CoTTA~\cite{wang2022continual} & 85.0 & 83.3 & 87.7 & 86.3 & 85.8 & 90.3 & 89.2 & 92.4 & 90.5 & 91.2 & 88.2 & 4.2 & 238.1 \\
 & EATA(C)~\cite{niu2022efficient} & 81.8 & 79.5 & 84.2 & 83.7 & 82.3 & 86.0 & 86.1 & 87.9 & 88.6 & 86.9 & 84.7 & 35.3 & 28.3 \\
 & EATA(O)~\cite{niu2022efficient} & 85.2 & 83.0 & 87.5 & 86.6 & 86.1 & 90.1 & 89.8 & 92.2 & 91.8 & 91.7 & 88.4 & 27.4 & 36.5 \\
 & EcoTTA(C)~\cite{song2023ecotta} & 84.6 & 80.9 & 85.5 & 84.7 & 84.3 & 88.0 & 86.9 & 91.0 & 89.8 & 89.5 & 86.5 & 31.2 & 32.1 \\
 & EcoTTA(O)~\cite{song2023ecotta} & 85.4 & \textbf{84.9} & 88.3 & 87.4 & 86.5 & 91.7 & 90.7 & 92.9 & 91.9 & 92.3 & 89.2 & 23.8 & 42.0 \\
 & MECTA~\cite{hong2023mecta} & 86.6 & 83.6 & 88.4 & 88.2 & 87.0 & \textbf{92.1} & \textbf{91.0} & 93.6 & 91.7 & 92.7 & 89.5 & 31.2 & 32.1 \\ \cline{2-15} 
 & Melon~\cite{wang2022melon} & 72.0 & 51.4 & 47.0 & 40.3 & 26.1 & 23.4 & 13.9 & 9.4 & 9.1 & 5.4 & 29.8 & 22.0 & 45.5 \\
 & TTE~\cite{lin2022device} & 68.7 & 65.8 & 71.5 & 70 & 68.9 & 73.8 & 72.2 & 74.7 & 74 & 75.5 & 71.5 & 29.5 & 33.9 \\
 & ElasticTrainer~\cite{huang2023elastictrainer} & 72.3 & 69.9 & 73.7 & 74.2 & 72.3 & 76.3 & 77 & 78.3 & 78.5 & 78.5 & 75.1 & 57.6 & 17.4 \\ \cline{2-15} 
\multirow{-11}{*}{\textbf{NICO++}} & \cellcolor[HTML]{EFEFEF}\sysname & \cellcolor[HTML]{EFEFEF}\textbf{86.9} & \cellcolor[HTML]{EFEFEF}84.8 & \cellcolor[HTML]{EFEFEF}\textbf{89.0} & \cellcolor[HTML]{EFEFEF}\textbf{88.5} & \cellcolor[HTML]{EFEFEF}\textbf{87.7} & \cellcolor[HTML]{EFEFEF}91.6 & \cellcolor[HTML]{EFEFEF}\textbf{91.0} & \cellcolor[HTML]{EFEFEF}\textbf{93.7} & \cellcolor[HTML]{EFEFEF}\textbf{92.9} & \cellcolor[HTML]{EFEFEF}\textbf{92.9} & \cellcolor[HTML]{EFEFEF}\textbf{89.9} & \cellcolor[HTML]{EFEFEF}\textbf{76.4} & \cellcolor[HTML]{EFEFEF}\textbf{13.1} \\ \hline
\end{tabular}
}
\label{exp:table_1}
\end{table*}

\begin{table*}[t]
\centering
\footnotesize
\caption{Test accuracy (\%), adaptation throughput (fps), and latency (ms) with continual corruption shifts.}
\renewcommand{\arraystretch}{1.25}
\scalebox{0.84}{
\begin{tabular}{cccccccccccccccccccc}
\hline
\multicolumn{2}{c}{\textbf{Time}} & \multicolumn{15}{c}{\textbf{t —>}} &  &  &  \\ \cline{1-17}
\textbf{Dataset} & \textbf{Method} & \textbf{Gaus.} & \textbf{Shot} & \textbf{Impu.} & \textbf{Defo.} & \textbf{Glas.} & \textbf{Moti.} & \textbf{Zoom} & \textbf{Snow} & \textbf{Fros.} & \textbf{Fog} & \textbf{Brig.} & \textbf{Cont.} & \textbf{Elas.} & \textbf{Pixe.} & \textbf{Jpeg} & \multirow{-2}{*}{\begin{tabular}[c]{@{}c@{}} \textbf{Mean} \\ \textbf{ACC.(\%)} \end{tabular}} & \multirow{-2}{*}{\begin{tabular}[c]{@{}c@{}} \textbf{Th.} \\ \textbf{(fps)} \end{tabular}} & \multirow{-2}{*}{\begin{tabular}[c]{@{}c@{}} \textbf{Latency} \\ \textbf{(ms)} \end{tabular}} \\ \hline
 & TENT~\cite{wang2021tent} & 73.1 & 75.2 & 67.8 & 79.6 & 68.6 & 81.2 & 85.6 & 78.0 & 87.1 & 86.5 & 81.8 & \textbf{90.3} & 69.5 & 77.4 & \textbf{77.3} & 78.6 & 29.2 & 34.3 \\
 & EATA~\cite{niu2022efficient} & 69.4 & 71.5 & 71.6 & 83.5 & 70.6 & 81.8 & 85.9 & 79.5 & 83.1 & 86.1 & 87.1 & 86.8 & 75.1 & 83.1 & 73.1 & 79.2 & 27.5 & 36.4 \\
 & EcoTTA~\cite{song2023ecotta} & 65.7 & 68.7 & 70.6 & 78.1 & 68.1 & 79.7 & \textbf{87.9} & 77.3 & 85.8 & 89.3 & 90.2 & 82.3 & 72.9 & 77.4 & 72.9 & 77.8 & 23.8 & 42.1 \\
 & MECTA~\cite{hong2023mecta} & \textbf{74.7} & 70.1 & 70.4 & 83.2 & 68.5 & \textbf{84.9} & 83.3 & 79.2 & 85.2 & \textbf{90.9} & 91.9 & 85.2 & 76.3 & 85.7 & 72.2 & 80.1 & 31.0 & 32.3 \\
 & BFTT3D\cite{wang2024backpropagation} & 44.5 & 45.3 & 51.2 & 61.9 & 41.7 & 54.9 & 66.0 & 56.4 & 60.0 & 65.7 & 65.3 & 57.9 & 51.9 & 58.9 & 47.9 & 55.3 & 163.9 & 6.1 \\
 & LAME\cite{boudiaf2022parameter} & 47.8 & 43.4 & 50.6 & 59.0 & 44.9 & 55.4 & 67.4 & 56.6 & 62.1 & 58.5 & 62.6 & 68.0 & 55.7 & 62.6 & 45.3 & 56.0 & 175.4 & 5.7 \\
 & OFTTA\cite{wang2024optimization} & 48.6 & 46.1 & 46.9 & 57.9 & 47.0 & 64.6 & 63.3 & 54.5 & 57.6 & 67.6 & 66.9 & 66.4 & 53.9 & 61.6 & 50.6 & 56.9 & \textbf{192.3} & \textbf{5.2} \\ \cline{2-20}
 & ElasticTrainer~\cite{huang2023elastictrainer} & 53.6 & 59.9 & 59.8 & 72.8 & 57.9 & 70.4 & 74.3 & 68.7 & 71.5 & 69.2 & 73.7 & 75.8 & 61.4 & 67.2 & 60.5 & 66.4 & 58.6 & 17.1 \\ \cline{2-20}
\multirow{-9}{*}{\textbf{CIFAR-10C}} & \cellcolor[HTML]{EFEFEF}\textbf{\sysname} & \cellcolor[HTML]{EFEFEF}69.2 & \cellcolor[HTML]{EFEFEF}\textbf{75.4} & \cellcolor[HTML]{EFEFEF}\textbf{72.6} & \cellcolor[HTML]{EFEFEF}\textbf{83.6} & \cellcolor[HTML]{EFEFEF}\textbf{71.7} & \cellcolor[HTML]{EFEFEF}79.6 & \cellcolor[HTML]{EFEFEF}84.7 & \cellcolor[HTML]{EFEFEF}\textbf{85.2} & \cellcolor[HTML]{EFEFEF}\textbf{87.4} & \cellcolor[HTML]{EFEFEF}85.8 & \cellcolor[HTML]{EFEFEF}\textbf{92.6} & \cellcolor[HTML]{EFEFEF}84.8 & \cellcolor[HTML]{EFEFEF}\textbf{76.6} & \cellcolor[HTML]{EFEFEF}\textbf{85.9} & \cellcolor[HTML]{EFEFEF}74.8 & \cellcolor[HTML]{EFEFEF}\textbf{80.7} & \cellcolor[HTML]{EFEFEF}79.1 & \cellcolor[HTML]{EFEFEF}12.6 \\ \hline
\end{tabular}
}
\label{exp:table_2}
\end{table*}


\begin{table}[t]
\centering
\footnotesize
\caption{Performance comparison on CIFAR100-C.}
\renewcommand{\arraystretch}{1.1}
\scalebox{1}{
\begin{tabular}{ccccc}
\hline
\multicolumn{2}{c}{\textbf{Time}} &  &  &  \\ \cline{1-2}
\textbf{Dataset} & \textbf{Method} & \multirow{-2}{*}{\begin{tabular}[c]{@{}c@{}} \textbf{Mean} \\ \textbf{ACC.(\%)} \end{tabular}} & \multirow{-2}{*}{\textbf{Th.(fps)}} & \multirow{-2}{*}{\textbf{Latency(ms)}} \\ \hline
 & TENT~\cite{wang2021tent} & 60.1 & 26.1 & 38.3 \\
 & EATA~\cite{niu2022efficient} & 58.8 & 24.3 & 41.2 \\
 & EcoTTA~\cite{song2023ecotta} & 59.3 & 21.5 & 46.6 \\
 & MECTA~\cite{hong2023mecta} & 61.8 & 27.2 & 36.8 \\
 & BFTT3D\cite{wang2024backpropagation} & 44.1 & 158.7 & 6.3 \\
 & LAME\cite{boudiaf2022parameter} & 41.8 & 163.9 & 6.1 \\
 & OFTTA\cite{wang2024optimization} & 42.5 & \textbf{175.4} & \textbf{5.7} \\ \cline{2-5} 
 & ElasticTrainer~\cite{huang2023elastictrainer} & 49.1 & 56.9 & 17.6 \\ \cline{2-5} 
\multirow{-9}{*}{\textbf{CIFAR100-C}} & \cellcolor[HTML]{EFEFEF}\sysname & \cellcolor[HTML]{EFEFEF}\textbf{62.2} & \cellcolor[HTML]{EFEFEF}76.2 & \cellcolor[HTML]{EFEFEF}13.1 \\ \hline
\end{tabular}
}
\label{exp:table_cifar100c}
\end{table}

\subsection{Main Results}
\subsubsection{Accuracy vs. Latency}
\label{Accuracy_vs_Latency}
This experiment evaluates performance during adaptation to two types of \textit{continual shifts}. 
We pre-train a ResNet50 on the "outdoor" sub-dataset in NICO++ and sequentially adapt it to five other sub-datasets with \textit{environmental} shifts (autumn, dim, grass, rock, and water) on $D_1$ (see \tabref{exp:table_1}). 
\sysname achieves a better balance between adaptation accuracy and latency, achieving $2\times \sim 3.5\times$ lower latency than Tent, EATA, and EcoTTA, and up to $18.2\times$ lower than CoTTA.
\lsc{
While ElasticTrainer achieves a $2\times$ speedup over other baselines, \sysname's latency is only $75.3\%$ of ElasticTrainer's. 
Additionally, \sysname outperforms existing TTA methods in parallel inference/adaptation, achieving the highest adaptation accuracy with up to a $5.2\%$ improvement over generic and efficient TTA baselines.
MECTA offers comparable accuracy due to its adaptive BN layers. 
In label-free environments, \sysname surpasses Melon by $60.1\%$, effectively reducing overfitting through sparse updating. 
It also shows accuracy gains of $18.4\%$ over TTE and $14.8\%$ over ElasticTrainer, thanks to its backpropagation-free design and timely layer importance assessment. 
In contrast, TTE and ElasticTrainer's reliance on backpropagation for layer importance evaluation results in significant performance drops when labels are unavailable.
}

We further compare \sysname with baselines under \textit{corruption} shifts, another common in mobile applications due to sensor degradation. 
\sysname maintains the best accuracy-latency balance while adapting to corrupted CIFAR10 (see Tab. \ref{exp:table_2}) and CIFAR100 (see Tab. \ref{exp:table_cifar100c}) datasets on device \(D1\).
In CIFAR10-C shifts, \sysname outperforms EcoTTA, MECTA, and ElasticTrainer, achieving accuracy improvements of 2.9\%, 0.6\%, and 14.3\%, along with accelerations of 3.3$\times$ 2.6$\times$ and 1.2$\times$, respectively. 
For CIFAR100-C, \sysname surpasses these baselines with similar accuracy gains and accelerations.
We also compare \sysname with backward-free TTA methods (BFTT3D, LAME, and OFTTA) on CIFAR10-C and CIFAR100-C (see Tab. \ref{exp:table_2}). 
Although \sysname is on average 2.17 times slower, backward-free methods experience a 18.1\%-25.4\% accuracy drop after more than two continual shifts, which is unacceptable for mobile tasks. 
\lsc{These methods perform adequately in single shift scenarios but lack adaptability during continual shifts and significant image changes because prohibiting backward and parameter updating limit the model's optimization capacity\cite{yu2023benchmarking}.
OFTTA and BFTT3D address prior distribution shifts but provide minimal improvement for appearance shifts, while LAME performs worse than a non-adaptive model in scenarios with sensor degradation, such as \textit{Shot}, \textit{Gaus.}, and \textit{Pixe.}.}

In summary, \sysname outperforms other baselines in adaptation latency and accuracy across diverse non-stationary shifts due to its efficient sparse updating with accurate online predictors for unit update importance and latency.

\begin{table}[t]
\centering
\footnotesize
\caption{Adaptation latency on diverse devices.}
\renewcommand{\arraystretch}{1.1}
\scalebox{1.05}{
\begin{tabular}{|c|ccc|ccc|}
\hline
                & \multicolumn{3}{c|}{NICO++}                                      & \multicolumn{3}{c|}{CIFAR10-C}                                    \\ \hline
Diverse devices & \multicolumn{1}{c|}{$D_1$} & \multicolumn{1}{c|}{$D_2$} & $D_3$  & \multicolumn{1}{c|}{$D_1$} & \multicolumn{1}{c|}{$D_2$} & $D_3$  \\ \hline
EATA \cite{niu2022efficient} & \multicolumn{1}{c|}{39.5}  & \multicolumn{1}{c|}{318.5} & 1960.8 & \multicolumn{1}{c|}{36.4}  & \multicolumn{1}{c|}{344.8} & 2325.6 \\ \hline
EcoTTA \cite{song2023ecotta} & \multicolumn{1}{c|}{44.1}  & \multicolumn{1}{c|}{322.6} & 2234.7 & \multicolumn{1}{c|}{43.1}  & \multicolumn{1}{c|}{405.2} & 3031.8 \\ \hline
ElasticTrainer \cite{huang2023elastictrainer} & \multicolumn{1}{c|}{16.4}  & \multicolumn{1}{c|}{144.9} & 831.6  & \multicolumn{1}{c|}{16.2}  & \multicolumn{1}{c|}{136.6} & 1013.2 \\ \hline
Ours            & \multicolumn{1}{c|}{\textbf{12.9}}  & \multicolumn{1}{c|}{\textbf{113.6}} & \textbf{769.2}  & \multicolumn{1}{c|}{\textbf{12.5}}  & \multicolumn{1}{c|}{\textbf{121.9}} & \textbf{909.1}  \\ \hline
\end{tabular}
}
\label{exp:table_3}
\end{table}

\begin{figure}[t]
  \centering
  \includegraphics[width=0.49\textwidth]{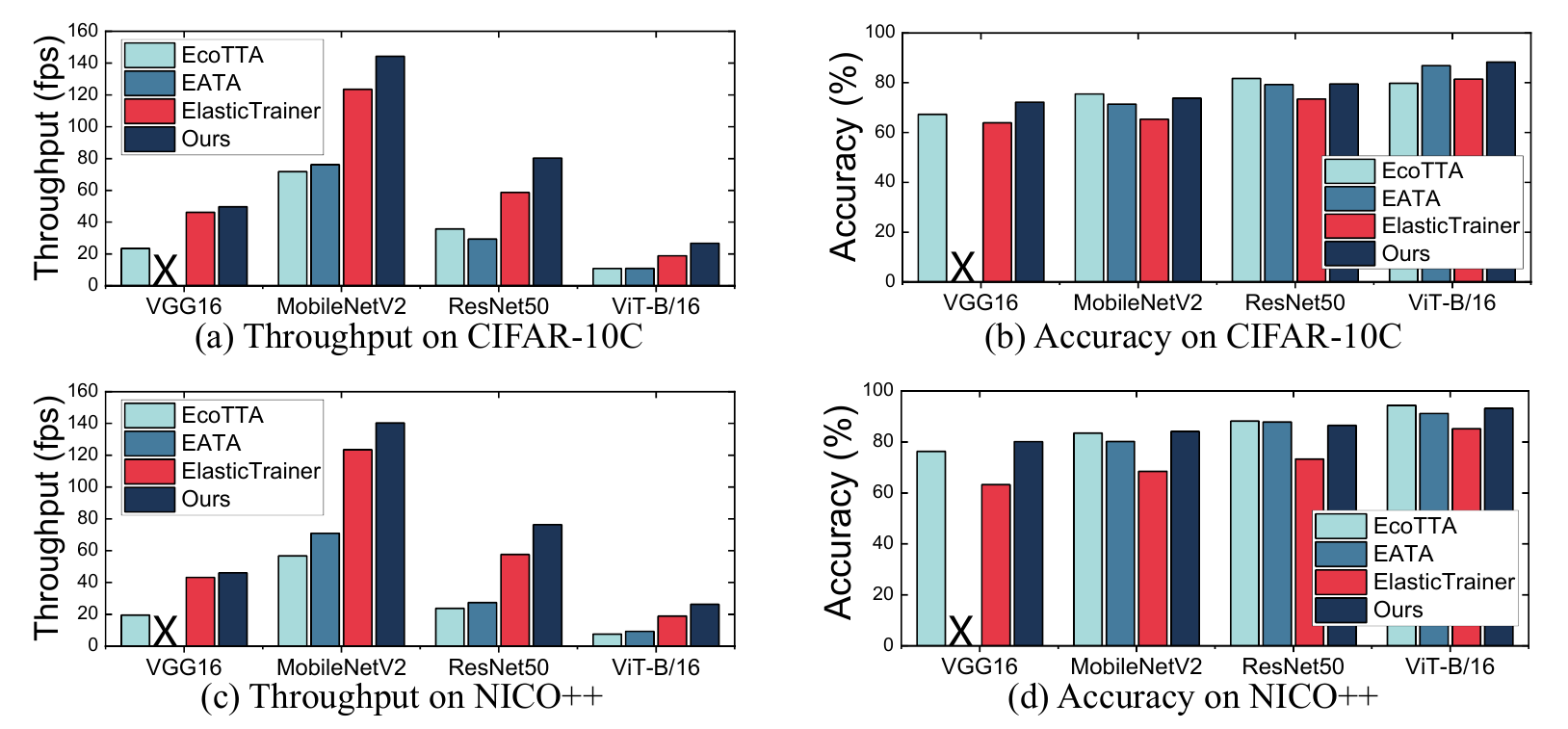}
   \caption{Adaptation performance with diverse DNNs.}
  \label{fig:Model_Architectures}
\end{figure}

\subsubsection{Performance across Diverse Devices}
\label{exp_GPUdevice}
This experiment verifies \sysname's effectiveness under different hardware and DVFS strategies by comparing to EATA, EcoTTA, and ElasticTrainer on three devices (\ie $D_1 \sim D_3$) using NICO++ and CIFAR10-C shifted data. 
As shown in \tabref{exp:table_3}, \sysname achieves the lowest adaptation latency, being at most 32.7\%, 29.3\%, and 77.1\% that of EATA, EcoTTA, and ElasticTrainer, respectively.
The latency difference between \sysname and ElasticTrainer is smaller on device $D_3$ due to increased GPU core parallelization, increasing differences between runtime and offline latency, highlighting \sysname's advantage.

\subsubsection{Performance using Diverse Model Architectures}
We compare \sysname's adaptation accuracy and throughput with EATA, EcoTTA, and ElasticTrainer on ViT-B/16, VGG16, MobileNetV2, and ResNet50 using the NICO++ and CIFAR10-C datasets on device $D1$. 
As shown in \figref{fig:Model_Architectures}, \sysname achieves $2\times \sim 3\times$ acceleration while maintaining high accuracy across all models. 
ElasticTrainer struggles with low accuracy in unsupervised conditions relying on backpropagation. 
MobileNetV2's simplified structure results in inferior learnability and adaptability, requiring more epochs to converge. 
EcoTTA and EATA fall short for lightweight DNNs due to fixed sparse updating. 
In contrast, \sysname flexibly selects layers to reduce accuracy gaps and employs a unit-based latency predictor to ensure an accuracy-latency balance, even on transformer architectures like ViT-B/16.

\subsection{Micro-benchmarks}

\begin{figure}[t]
  \centering
  \includegraphics[width=0.45\textwidth]{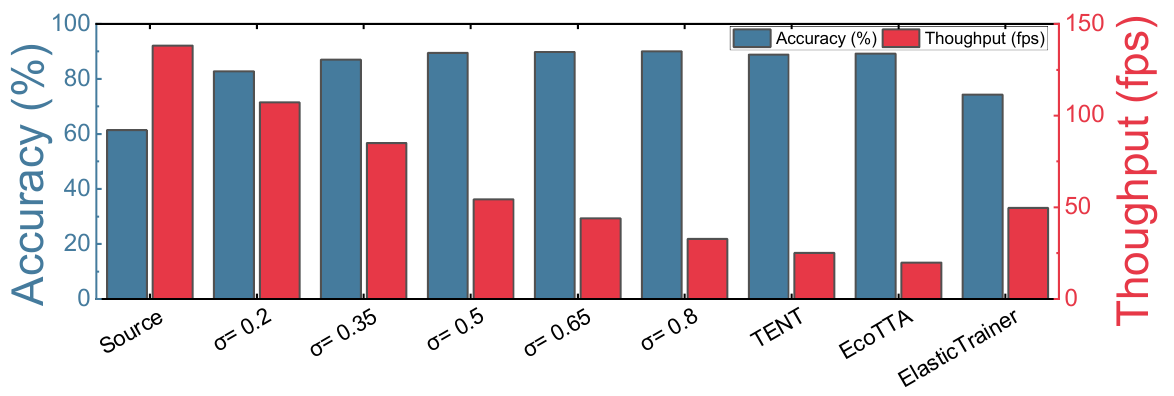}
  \vspace{-2mm}
  \caption{Impact of acceleration factor.}
  \vspace{-3mm}
  \label{fig:speed_up_objective}
\end{figure}

\subsubsection{Impact of Expected Acceleration Factor $\sigma$}
This experiment tests the impact of the expected acceleration factor $\sigma$ in \equref{search_target} on \sysname. 
From \figref{fig:speed_up_objective}, \sysname achieves relatively balanced latency and accuracy when $\sigma$ varies from $0.2$ to $0.5$.
When $\sigma$ is larger than $0.5$, the accuracy saturates.
When $\sigma$ is lower than $0.2$ \sysname shows notable accuracy loss.
We empirically set $\sigma=0.33$ in \sysname.


\subsubsection{Impact of Adaptation Rate}
\begin{figure}[t]
	\centering 
	\subfloat[]{\label{fig:exp_alpha}
		\includegraphics[height=0.31\linewidth]{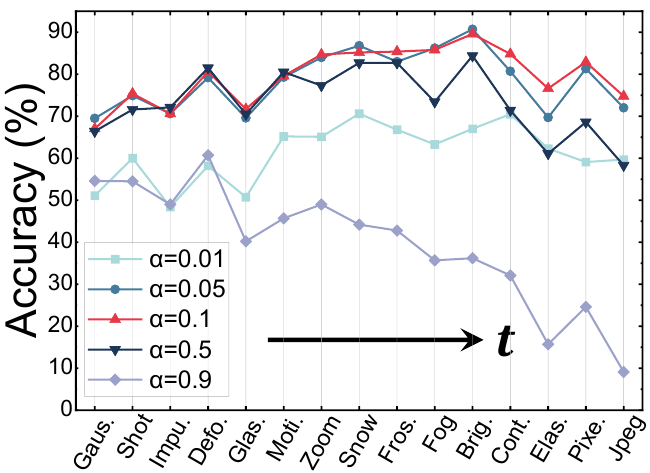}}
    \hspace{7mm}
    \subfloat[]{\label{fig:exp_lr}
		\includegraphics[height=0.31\linewidth]{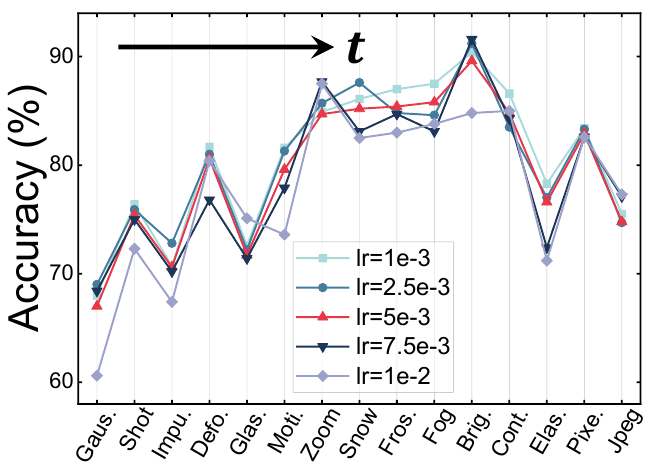}}
\caption{Impact of (a) $\sigma$ and (b) $lr$.}
\label{fig:exp_alpha_lr}
\end{figure}

We test the \sysname's sensitivity to adaptation rate using  CIFAR10-C and ResNet50, including integration rate $\alpha$ and learning rate $lr$.
\sysname shows robustness to $lr$ variations (see Fig. \ref{fig:exp_lr}) but is sensitive to $\alpha$. 
A small $\alpha$ leads to the model hardly learning new scenes, while a large $\alpha$ causes the model to easily overfit to new scenes (see Fig. \ref{fig:exp_alpha}).
Therefore, we empirically set $\alpha=0.1$ in \sysname and $lr$=5$e$-3 by default.

\begin{table}[t]
\centering
\footnotesize
\caption{Impact of batch sizes (BS).}
\renewcommand{\arraystretch}{1.2}
\scalebox{1.1}{
\begin{tabular}{|c|c|c|c|c|}
\hline
                & BS=1 & BS=2 & BS=4 & BS=16 \\ \hline
Accuracy (\%)    & 79.5 & 84.5 & 88.5 & 89.7  \\ \hline
Throughput (fps) & 19.2 & 43.2 & 79.6 & 340.2 \\ \hline
Peak memory (MB) & 2108 & 2210 & 2462 & 3794  \\ \hline
\end{tabular}
}
\label{exp:table_4}
\end{table}

\lsc{
\subsubsection{Impact of Batch Sizes}
\label{Impact_of_Batch_Sizes}
For stay responsive to non-stationary environments, effective adaptation on small batch sizes is expected.
\tabref{exp:table_4} shows the performance of \sysname with batch sizes of 1, 2, 4, and 16 with NICO++ on device $D_1$. 
Even with a batch size of 1, Adashadow keeps a high accuracy with $79.5\%$, supporting dynamic batch size adjustments to maintain responsiveness. 
As the batch size increases, throughput and accuracy improve at the cost of memory usage. 
With a batch size of 16, it surpasses the device's memory budget.
%
Thus, we set $batch\ size=4$ with $79.2fps$ throughput by default in \sysname. 
}

    

\begin{figure}[t]
	\centering 
	\subfloat[]{\label{exp:table_adaptation_mode}
		\includegraphics[height=0.31\linewidth]{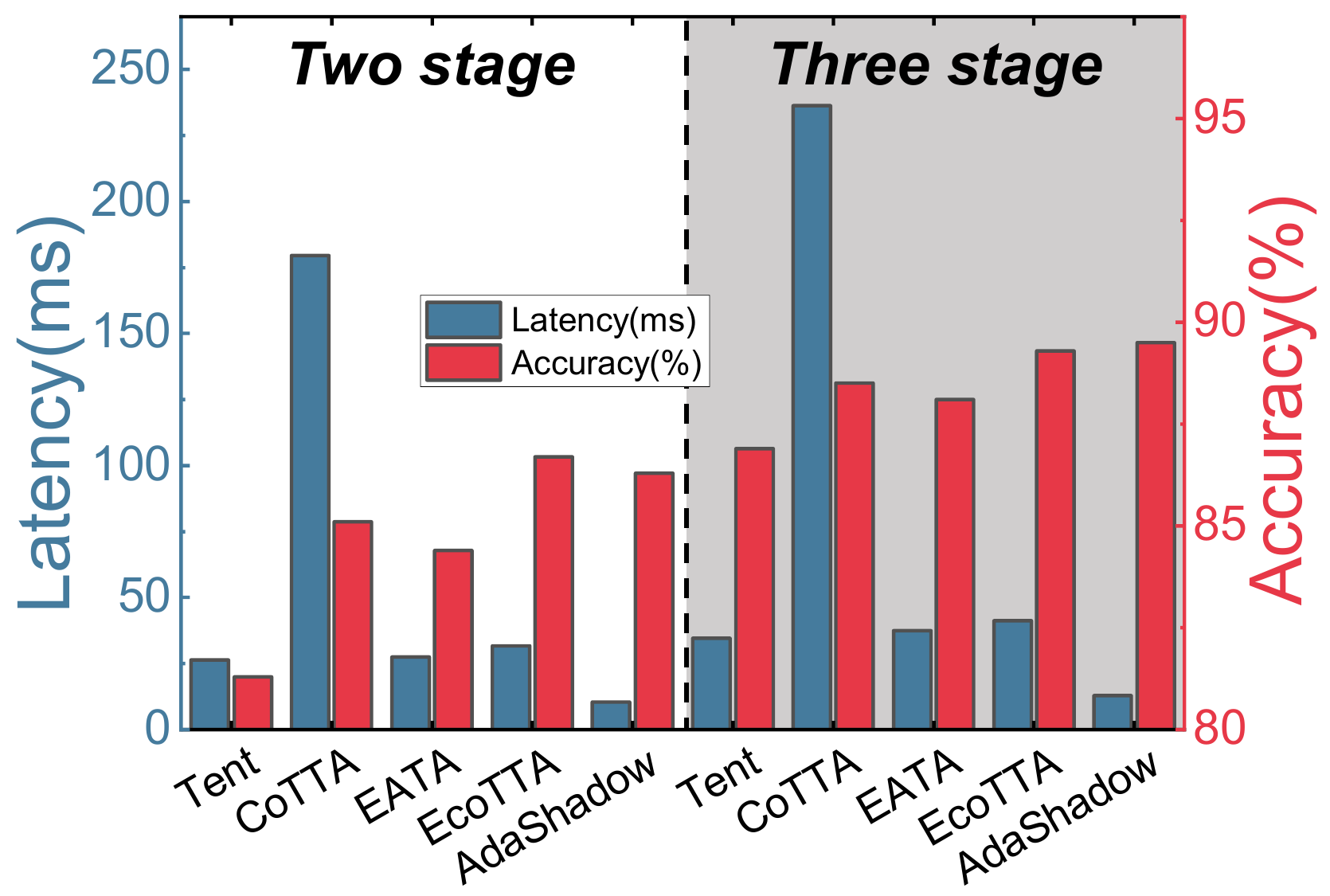}}
    \hspace{5mm}
    \subfloat[]{\label{exp:table_9}
		\includegraphics[height=0.31\linewidth]{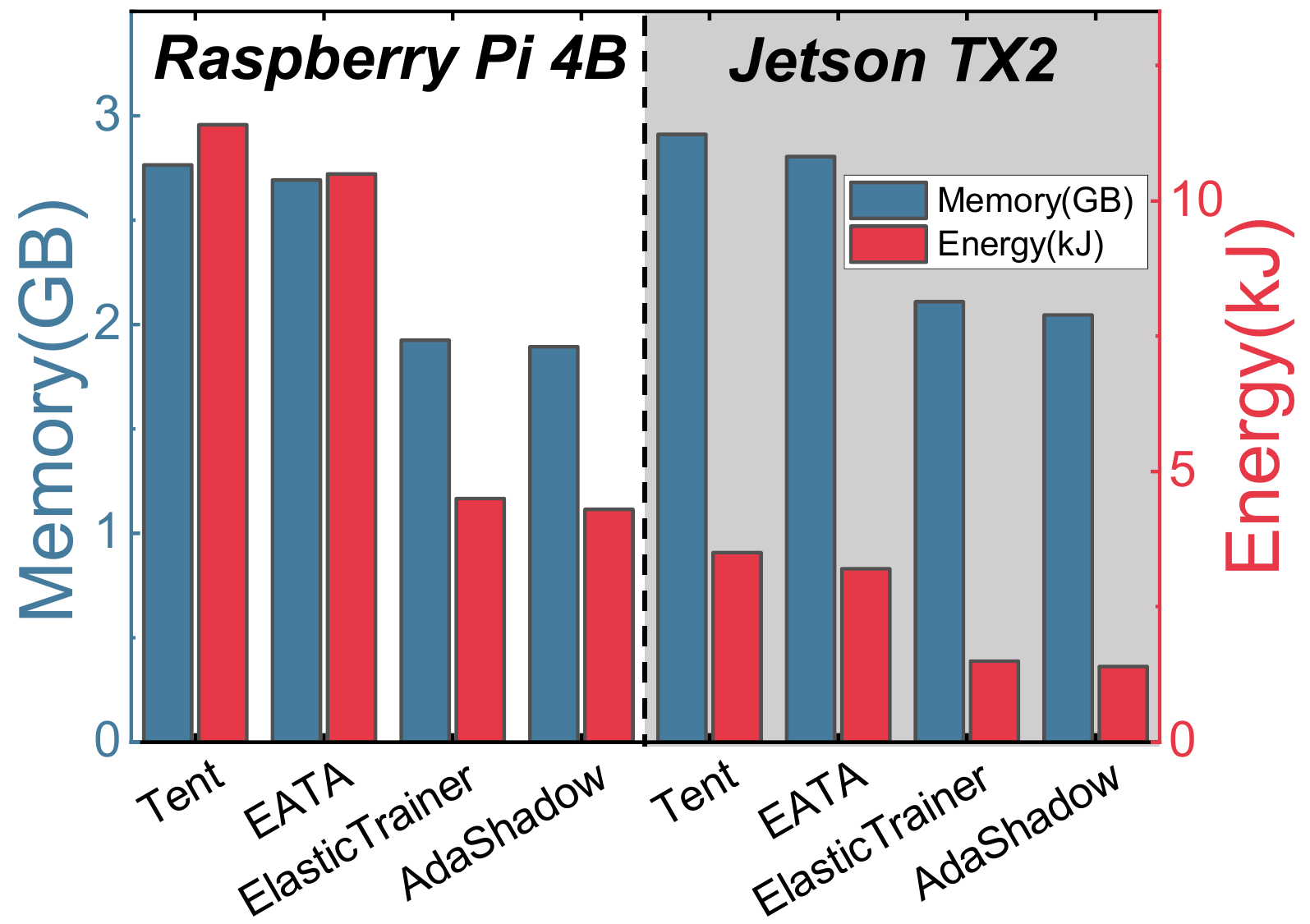}}
\caption{Evaluation of (a) different adaptation modes and (b) system overhead.}
\label{fig:exp_alpha_lr}
\end{figure}

\subsubsection{Generalizing to Two-stage Adaptation Mode.}
This experiment tests the \sysname's efficiency of different adaptation modes using NICO++ and ResNet50.
We compare the effectiveness of \sysname in three-stage (forward-backward-reforward) and two-stage (forward-backward) online TTA settings ( \tabref{exp:table_adaptation_mode}).
Results show that \sysname also aligns well with the traditional two-stage (forward-backward) mode, achieving significant acceleration by optimizing the sparse updating in the backward pass.

\subsubsection{Performance of Latency Predictor}
We test \sysname's latency predictor under offline and $dr_1 \sim dr_4$ conditions (the same settings mentioned in \label{sec:design: profiler}) using  15,000 self-collected records from different systems, devices, and models.
\lsc{
We run and measure the network's \textit{offline latency} in a stable environment (25°C, with no competing processes for computation or memory). 
In the dynamic online environment, after predicting execution latency, the network is executed to collect \textit{actual latency} immediately.
}
Considering dynamic kernel and memory resources, the average prediction error between \sysname's latency predictor and ground truth under $dr_1 \sim dr_4$ are $\leq$ 3.3\% (see  \tabref{exp:latency_predictor_accuracy}), showing high accuracy and stability.


\begin{table}[t]
\centering
\footnotesize
\caption{Error of factors $\pi_1$ and $\pi_2$.}
\renewcommand{\arraystretch}{1.1}
\scalebox{1.1}{
\begin{tabular}{|c|c|c|c|c|c|}
\hline
 & \textbf{Offline} & \textbf{$dr_1$} & \textbf{$dr_2$} & \textbf{$dr_3$} & \textbf{$dr_4$} \\ \hline
\textbf{$\pi_1$} & 1.0 & 1.6 & 4.3 & 1.0 & 7.0 \\ \hline
\textbf{$\pi_2$} & 1.0 & 1.0 & 1.0 & 2.4 & 2.4 \\ \hline
\textbf{Offline latency} & 45.8 & 45.8 & 45.8 & 45.8 & 45.8 \\ \hline
\textbf{Actual latency} & 45.3 & 73.5 & 187.2 & 50.5 & 309.9 \\ \hline
\textbf{Predicted latency} & 45.8 & 71.7 & 181.7 & 52.2 & 299.8 \\ \hline
\textbf{Error rate (\%)} & 1.2 & 2.4 & 2.9 & 3.3 & 3.2 \\ \hline
\end{tabular}
}
\label{exp:latency_predictor_accuracy}
\end{table}

\subsubsection{System Overhead}
\sysname can efficiently tune memory usage by selective updates of crucial layers in a dynamic adaptive manner. 
In Table \ref{exp:table_9}, \sysname achieves a peak memory usage of $\leq 2045 MB$, lower than Tent, EATA, and ElasticTrainer.
Moreover, \sysname significantly decreases the frequency and duration of backpropagation, reducing 56\% and 60\% energy cost than Tent and EATA.

\subsection{Case Studies}
\label{sec:exp:case}

\begin{figure}[t]
    \centering 
    \subfloat[]{\label{fig:InferenceAfterAdaptation}
    \includegraphics[width=0.86\linewidth]{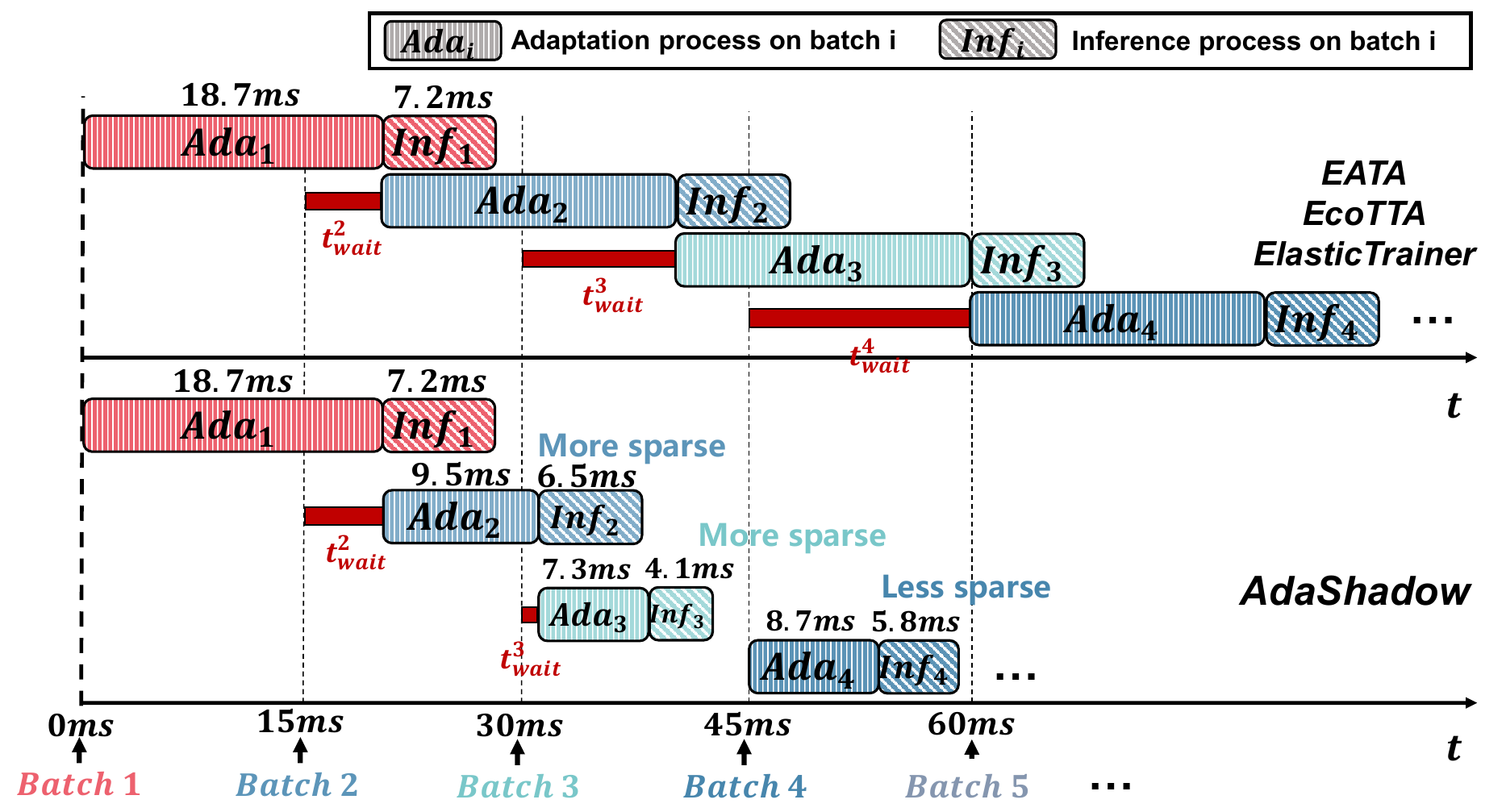}}
    \\
     \subfloat[]{\label{fig:InferenceImmediately}
    \includegraphics[width=0.88\linewidth]{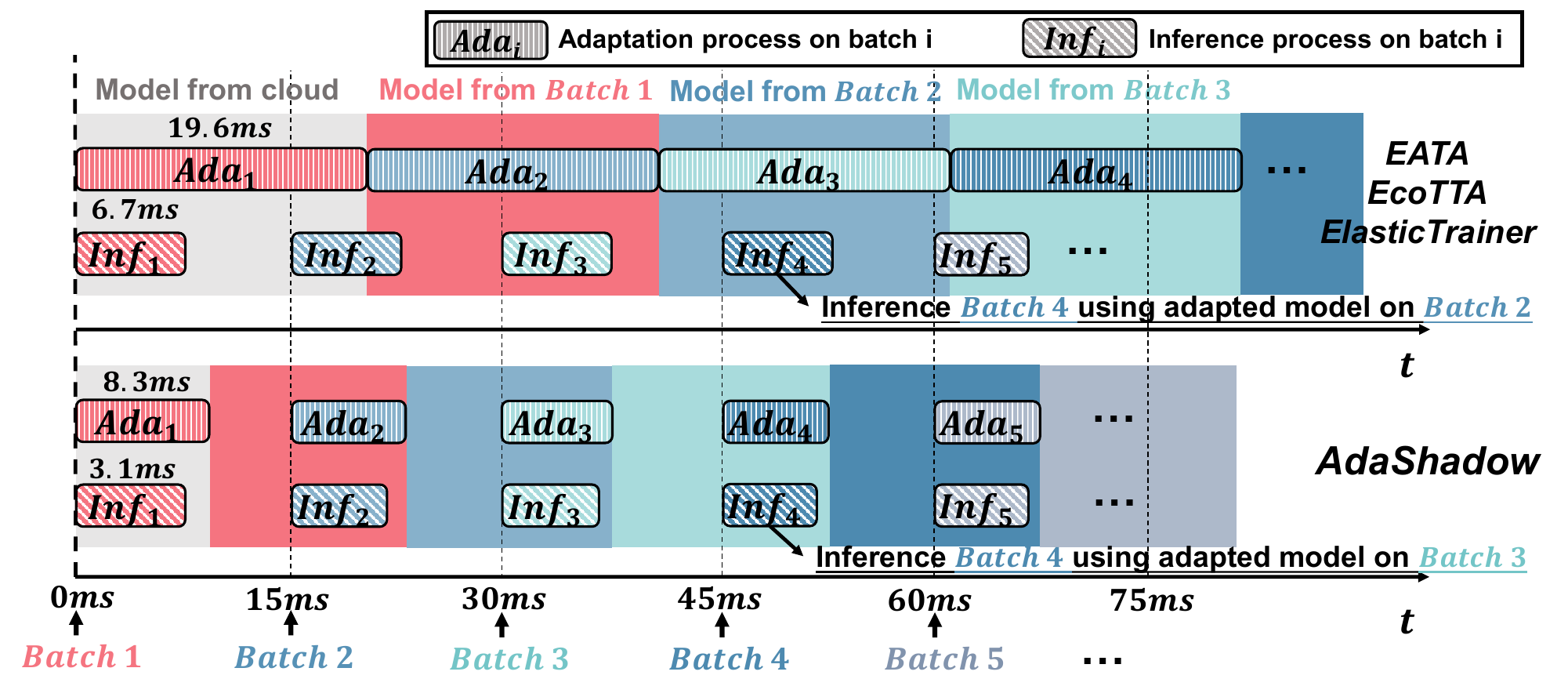}}
    \caption{Illustration of (a) sequential and (b) parallel adaptation/inference mode.}
\label{fig:ContinuousInference}
\end{figure}

\begin{table}[t]
\centering
\footnotesize
\caption{Performance comparison in sequential adaptation/inference mode.}
\renewcommand{\arraystretch}{1.1}
\scalebox{1.03}{
\begin{tabular}{|c|c|c|c|}
\hline
                & Accuracy (\%) & Throughput (fps) & Turnaround time \\ \hline
EATA \cite{niu2022efficient}          & 70.8     & 35.3            & 2.11                   \\ \hline
EcoTTA \cite{song2023ecotta}          & 71.4     & 22.8            & 2.42                   \\ \hline
ElasticTrainer \cite{huang2023elastictrainer}  & 60.7     & 70.6            & 1.47                   \\ \hline
AdaShadow & 76.5     & 80.8            & 1.37                   \\ \hline
\end{tabular}
}
\label{exp:table_5}
\end{table}

\subsubsection{Sequential Adaptation/Inference Mode}
This study showcases the usage of \sysname, where it first adapts the model upon a new batch of data before inference on the same batch using the updated model (see \figref{fig:InferenceAfterAdaptation}). 
That is, the inference on $batch 1$ is after the adaptation on $batch 1$, and the adaption on $batch 2$ also starts after the adaptation on $batch 1$.
This execution mode slightly prioritizes accuracy over latency and is widely used in applications such as AR and VR.
We set the time window to $15ms$, the batch size to $4$, and test with continuously played CIFAR10-C on device $D_1$.
As shown in \figref{fig:InferenceAfterAdaptation}, this execution mode introduces waiting time, making latency a potential bottleneck. 
We quantify such waiting time via the \textit{turnaround time} for batch $i$, \ie $r^i=\frac{T_{wait}^{i}+T_{a}^{i}+T_{re}^{i}}{T_{a}^{i}+T_{re}^{i}}$, a common metric in OS process scheduling, where $T_{wait}^{i}$ is the wait time of batch $i$.
We then report the average $r$ across all arriving data batches as the total waiting time.
Since \sysname dynamically adjusts the acceleration factor $\sigma$, it can gradually decrease $r$, where \sysname scales $\sigma$ to tune $t_{wait}^{2}$, $t_{wait}^{3}$, and $t_{wait}^{i}$ of subsequent arriving batch.
From \tabref{exp:table_5}, \sysname achieves higher inference accuracy with a lower $r$ than the baselines.

\begin{figure}[t]
  \centering
  \includegraphics[width=0.45\textwidth]{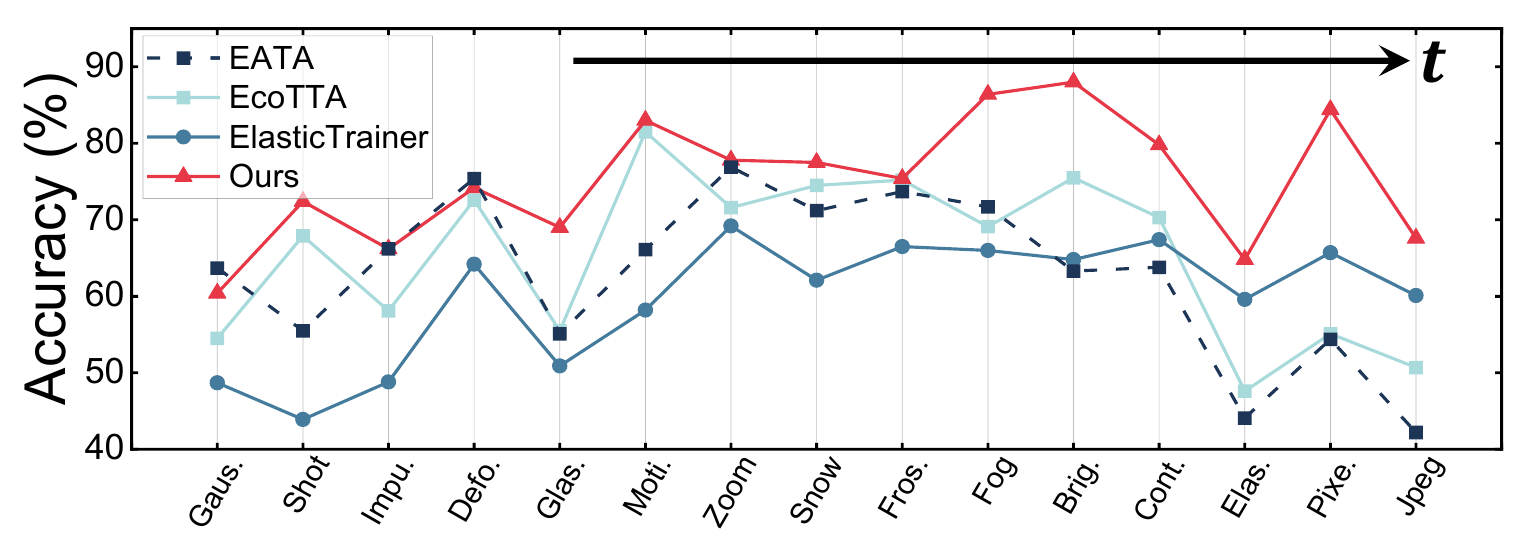}
  \caption{Accuracy comparison in parallel adaptation/inference mode.}
  \vspace{-4mm}
  \label{fig:Accuracy_Immediately}
\end{figure}

\begin{figure}[t]
  \centering  
  \includegraphics[width=0.48\textwidth]{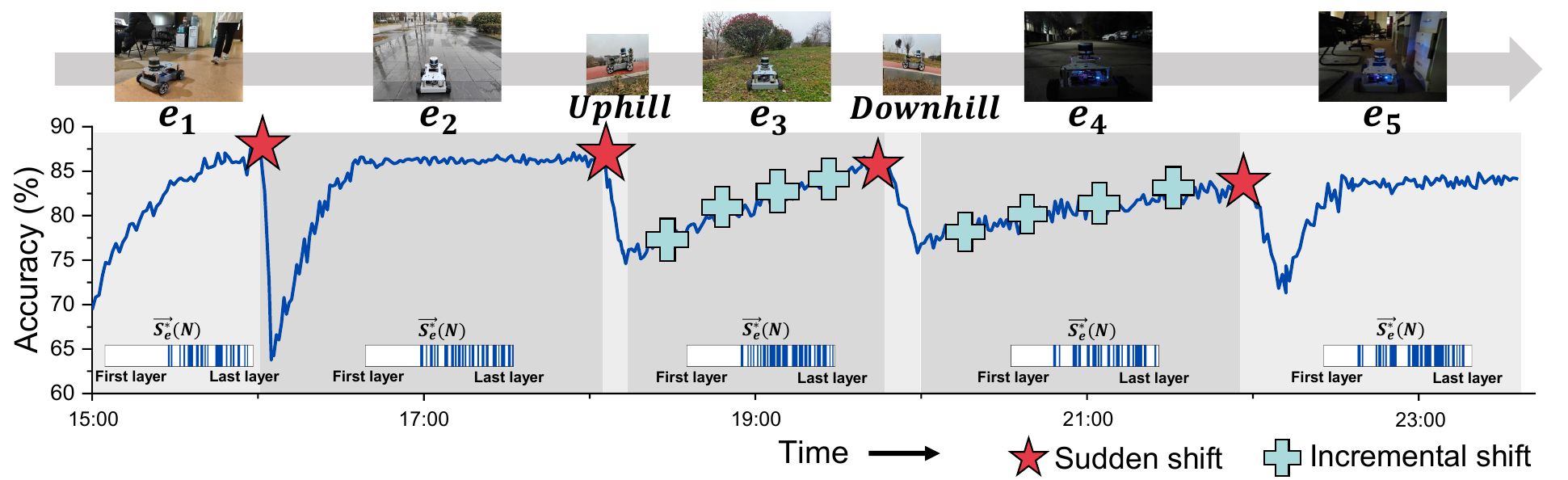}
  \caption{Case study illustration. $\vec{S}_{e}^{*}(N)$ is the corresponding optimal sparse updating strategy.}
  \vspace{-4mm}
  \label{fig:case_study}
\end{figure}

\subsubsection{Parallel Adaptation/Inference Mode}
This study demonstrates the usage of \sysname, where when a new batch of data arrives, it directly utilizes the current model for inference without waiting for the newly adapted one (see \figref{fig:InferenceImmediately}). 
This execution mode is preferred in real-time applications like drone object detection and autonomous driving.
We set the time window to $15ms$, the batch size to $4$, and test with CIFAR10-C on device $D_1$.
From \figref{fig:Accuracy_Immediately}, \sysname maintains the highest accuracy for its rapid and accurate model adaptation.
In contrast, other schemes fail to update model in time, leading to inaccurate inference using outdated models.

\subsubsection{Use Case}
This study evaluates \sysname in real-world mobile environments.
We built a vehicle platform to continuously collect video streams in a university campus in varying weather, illumination, and locations. 
We performed test-time DNN adaptation and inference during vehicle driving.
We summarized the environments encountered over time as: bright indoor ($e_1$) —\textgreater{} rainy road ($e_2$) —\textgreater{} uphill —\textgreater{} rainy grass ($e_3$) —\textgreater{} downhill —\textgreater{} night road ($e_4$) —\textgreater{} dark indoor ($e_5$).
From \figref{fig:case_study},  \sysname can promptly adapt to all these natural shifts with high accuracy and low latency by updating diverse layers. The blue bars below represent the layers that are selected for updating.

\section{Related Work}
\label{sec:related}

\fakeparagraph{Test-time Adaptation}
Test-time adaptation (TTA) is an emerging \textit{domain adaptation} setting with \textit{unlabeled} target data and assuming no access to any source data or supervision \cite{wang2021tent}.
Generic domain adaptation techniques \cite{ganin2015unsupervised, hoffman2018cycada, chen2019domain} are inherently supervised.
More recent label-free domain adaptation schemes \cite{wulfmeier2018incremental, taufique2022unsupervised} explored unsupervised adaption, yet still require assistance from the source data to assess and align data distributions. 
We refer readers to \cite{liang2023comprehensive} for a comprehensive review on domain adaptation.

Earlier TTA efforts focus on improving the adaptation \textit{accuracy} in this unsupervised, source-free setup.
Classical methods \cite{nado2020evaluating, schneider2020improving, wang2021tent} adapt the \textit{batch normalization (BN)} layers to the domain shifts by \textit{minimizing the entropy of the last layer's outputs on the training and testing data}.
Recent studies \cite{wang2022continual,song2023ecotta,chen2022contrastive} show that updating the BN layers alone might be insufficient for adaption to more drastic shifts and when the batch size is small and propose to also update parameters in the convolutional layers \cite{wang2022continual,chen2022contrastive} or adapters \cite{song2023ecotta}.
Since we aim at TTA with small batches, we also update all layers rather than restrict to BN layers.
Furthermore, we refine the adaption loss by incorporating outputs from all layers to enhance the robustness to small batches.

A few pioneer studies explore improving the \textit{efficiency} of TTA.
EATA \cite{niu2022efficient} focuses on data efficiency by only performing adaptation on important test samples.
EcoTTA \cite{song2023ecotta} adopts lightweight meta-networks attached to the original model for memory-efficient adaptation.
MECTA \cite{hong2023mecta} improves memory efficiency via an adaptive BN layer.
Yet, decreased computation or memory does not easily translate into reduced latency.
Additionally, \textit{retraining-free TTA}, including prototype-based approaches \cite{wang2024backpropagation,wang2024optimization} and consistency regularization \cite{boudiaf2022parameter}, enhance efficiency by eliminating the need for backpropagation. 
However, these approaches do not update model parameters,  limiting their ability to refine representations. 
This limitation leads to dramatic accuracy losses in mobile environments with continuous changes or drastic shifts~\cite{yu2023benchmarking} (evaluated in \secref{Accuracy_vs_Latency}), thereby motivating our work.


\fakeparagraph{On-device DNN Training}
Resource-efficient DNN training on mobile devices has attracted increasing research interest.
Common techniques include recomputation \cite{chen2016training, gim2022memory, wang2022melon, patil2022poet}, micro-batch \cite{huang2019gpipe, sohoni2019low, liu2022autopipe}, quantization \cite{xu2022mandheling, zhou2021octo, rastegari2016xnor}, sparse updating \cite{zaken2022bitfit, mudrakarta2019k, huang2023elastictrainer}, and memory swapping \cite{rhu2016vdnn, huang2020swapadvisor, chen2021cswap, wang2018superneurons}.
While memory swapping, recomputation, and micro-batch techniques can reduce memory usage, they cannot decrease or even increase training latency because of extra computations \cite{wang2022melon} or data transfers \cite{rhu2016vdnn}.
In contrast, sparse updating \cite{lin2022device, huang2023elastictrainer, zaken2022bitfit} often results in decreased latency as it reduces computations during backpropagation.

Sparse updating can be \textit{static} \cite{zaken2022bitfit, mudrakarta2019k, hu2022lora}, which updates fixed crucial DNN modules, or \textit{dynamic} \cite{wang2019e2,huang2023elastictrainer, hong2023mecta}, where the DNN modules to be updated are adjusted at runtime.
A static strategy is unfit for TTA since the modules critical for one domain may be less important for another.
Our work is most relevant to ElasticTrainer \cite{huang2023elastictrainer}, which also adopts a dynamic sparse updating strategy to accelerate DNN training.
However, ElasticTrainer \cite{huang2023elastictrainer} assumes labelled data, making it not directly applicable to TTA.
More importantly, it is primarily designed for offline training, where the training (adaptation) and inference are tightly coupled in TTA.
This coupling imposes stringent latency budget for layer importance assessment and necessitates awareness to resource dynamics for layer latency profiling, making ElasticTrainer \cite{huang2023elastictrainer} sub-optimal.
In contrast, \sysname achieves accurate and low-latency adaptation in mobile environments with fast-changing unlabelled data and highly dynamic resources.

\section{Conclusions}
This paper presents \sysname, a responsive test-time adaptation framework tailored for non-stationary mobile environments. 
It is the first endeavor toward near-real-time on-device DNN adaptation without source data or supervision. 
It accurately estimates layer importance in the forward pass without backpropagation, calibrates the offline layer latency measurements to runtime resource dynamics, and efficiently schedules optimal layer update strategy, overcoming TTA's latency bottleneck in mobile context without compromising accuracy.
Evaluations across diverse real-world scenarios and mobile devices show that \sysname achieves significant adaptation latency reductions and accuracy improvements. 
We plan to extend \sysname to mobile applications with stricter latency requirements and more complex DNNs. 
Integrating \sysname with federated learning, such as federated TTA, could enhance model robustness and user privacy.



\begin{acks}
This work was partially supported by the National Science Fund for Distinguished Young Scholars (62025205), the National Natural Science Foundation of China (No. 62032020,
6247074224, 62102317), and CityU APRC grant (No. 9610633).
\end{acks}

\bibliographystyle{ACM-Reference-Format}
\bibliography{adashadow-base}










\end{sloppypar}
\end{document}